\pdfoutput=1

\documentclass[11pt]{article}

\usepackage[preprint]{acl}

\usepackage{times}
\usepackage{latexsym}

\usepackage[T1]{fontenc}

\usepackage[utf8]{inputenc}

\usepackage{microtype}

\usepackage{inconsolata}

\usepackage{graphicx}
\usepackage{multirow}
\usepackage{booktabs}

\definecolor{navyblue}{rgb}{0.0, 0.0, 0.7}
\hypersetup{colorlinks, citecolor=navyblue, linkcolor=red}

\usepackage{amsmath,amssymb,amsfonts}
\DeclareMathOperator*{\argmax}{arg\,max}

\newcommand{\Autoref}[1]{%
  \begingroup%
  \def\algorithmautorefname{Algorithm}%
  \def\chapterautorefname{Chapter}%
  \def\sectionautorefname{Section}%
  \def\subsectionautorefname{Section}%
  \autoref{#1}%
  \endgroup%
}

\newtheorem{assumption}{Assumption}
\newtheorem{proposition}{Proposition}
\newtheorem{lemma}{Lemma}

\title{BAPO: Base-Anchored Preference Optimization for Overcoming Forgetting in Large Language Models Personalization}

\author{
Gihun Lee\textsuperscript{\rm 1}\quad
Minchan Jeong\textsuperscript{\rm 1}\quad
Yujin Kim\textsuperscript{\rm 1}\quad
Hojung Jung\textsuperscript{\rm 1} \quad \\
\textbf{Jaehoon Oh}\textsuperscript{\rm 2} \quad
\textbf{Sangmook Kim}\textsuperscript{\rm 3} \quad
\textbf{Se-Young Yun}\textsuperscript{\rm 1}
\\
\textsuperscript{\rm 1}Graduate School of AI, \, KAIST \;\,
\textsuperscript{\rm 2}Samsung Advanced Institute of Technology\\
\textsuperscript{\rm 3}Department of Electrical and Computer Engineering, UBC\\
}

\begin{document}
\maketitle
\begin{abstract}
While learning to align Large Language Models\,(LLMs) with human preferences has shown remarkable success, aligning these models to meet the diverse user preferences presents further challenges in preserving previous knowledge. This paper examines the impact of personalized preference optimization on LLMs, revealing that the extent of knowledge loss varies significantly with preference heterogeneity. Although previous approaches have utilized the KL constraint between the reference model and the policy model, we observe that they fail to maintain global knowledge and general alignment when facing personalized preferences. To this end, we introduce Base-Anchored Preference Optimization\,(BAPO), a simple yet effective approach that utilizes the initial responses of reference model to mitigate forgetting while accommodating personalized alignment. BAPO effectively adapts to diverse user preferences while minimally affecting global knowledge or general alignment.
\end{abstract}

\section{Introduction}

Large Language Models\,(LLMs)\,\citep{gpt4, llama2} have been successfully aligned with human preferences across various applications, ranging from summarization tasks to enhancing reasoning capabilities\,\citep{rlhf, learning_to_summarize_rlhf, zephyr, better_reasoners_alignment}. This alignment process involves collecting human feedback by presenting pairs of responses generated from the same user prompt and asking users to choose their preferred response\,\citep{hh_rlhf, ultrafeedback, rlaif, domain_specific_preference}. The LLMs learn from this preference data to produce responses that better match human preferences, effectively addressing the challenge of converting complex human expectations into tangible training objectives\,\citep{rlhf, alignment_survey, cpo_contrastive}. Known as preference optimization, this approach has become essential in the final stages of LLM training\,\citep{llama3, phi3, mixtral}.

\begin{figure*}[ht!]
    \centering
    \includegraphics[width=\textwidth]{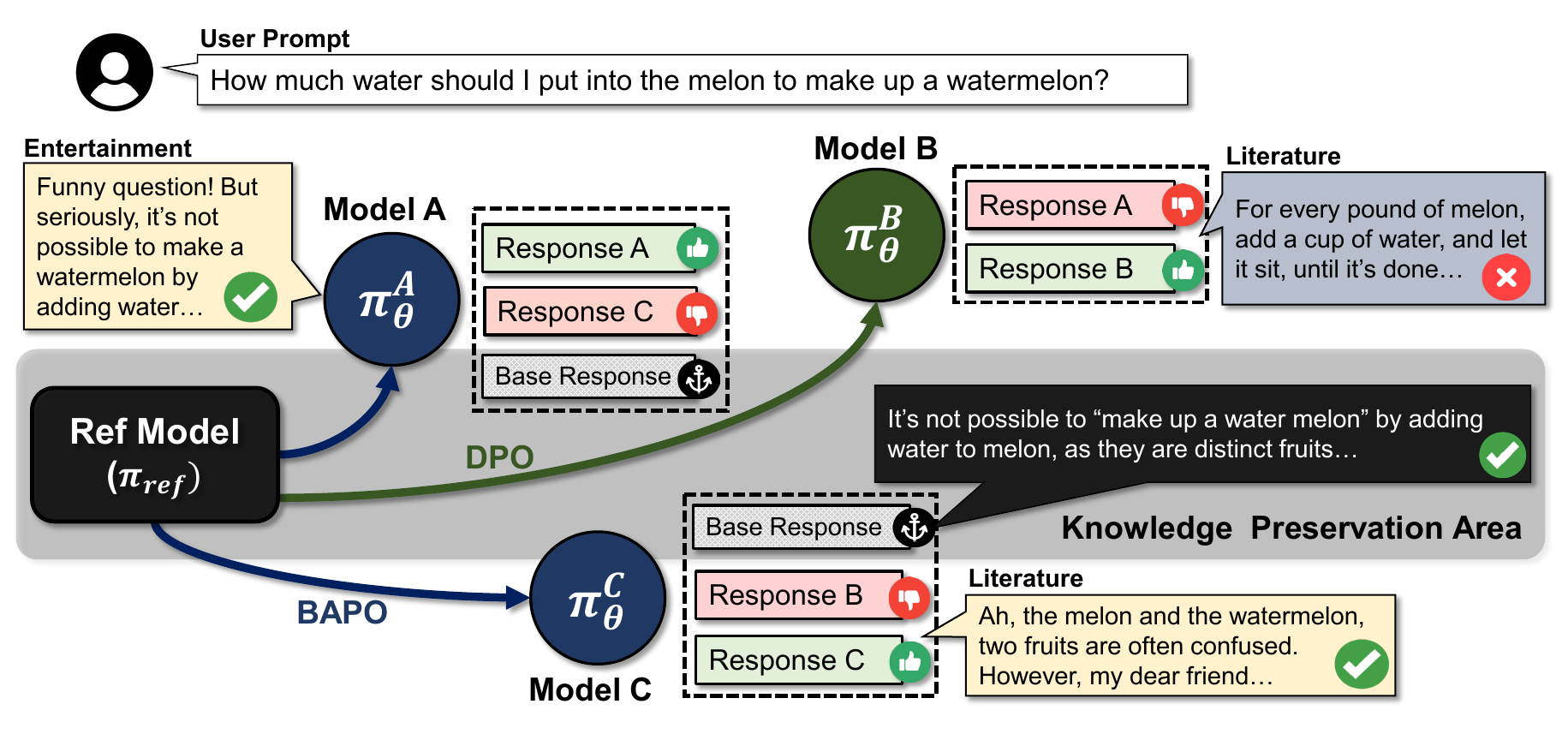}
    \vspace{-14pt}
    \caption{Overview of Base-Anchored Preference Optimization (BAPO): For a given user prompt, the base response achieves general alignment. \textbf{\textcolor{black!50!blue}{Models A}} and \textbf{\textcolor{black!50!blue}{Model C}}, fine-tuned with BAPO, maintain this alignment by anchoring to the base response. In contrast, \textbf{\textcolor{black!50!green}{Model B}}, fine-tuned with DPO, fails to preserve the knowledge from the base response, drifting away from the desired knowledge preservation area. The full example is provided in \autoref{appendix:main_full_example}.}
    \label{fig:main_bapo}
\end{figure*}

However, the common assumption in preference optimization is that all users share a uniform set of general preferences\,\citep{hh_rlhf, dpo, scerets_rlhf_ppo}, leading LLMs to align with an average of these preferences, as derived from collective feedback data\,\citep{morl_prompt, p_rlhf, cpo_controllable}. While effective for broadly accepted preferences like helpfulness and harmlessness, this approach does not account for the diversity of individual preferences in real-world scenarios\,\citep{psoups, more, domain_specific_preference, panacea}. For example, given the same context, one user might prefer a humorous response, while another might prefer a concise one. This reliance on averaged preferences often fails to capture the unique preferences of each user. This is known as the \textit{Condorcet Paradox}\,\citep{condorcet_paradox1, condorcet_paradox2} in social choice theory, where no single response consistently satisfies all users, leading to non-transitive preferences\,\citep{dpa_morl, nash_human_feedback}.

Recent studies have begun to tackle this challenge by fine-tuning instruction-tuned LLMs for personalized alignment\,\citep{psoups, more, reward_soups}. Although these personalized preference optimization approaches enable support for diverse user preferences\,\citep{p_rlhf, panacea, cpo_controllable}, the impact of learning to meet personalized preferences on previously acquired knowledge\,\citep{forecasting_forgetting_llms, online_merging_rlhf}, such as global knowledge\,\citep{loramoe} and general alignment\,\citep{mitigating_alignment_tax}, remains under\-explored.

In this work, we systematically analyze how personalizing LLMs according to diverse user preferences impacts their global knowledge and general alignment. Our findings reveal that the extent of knowledge loss heavily depends on these preferences, often inducing significant declines in specific areas of knowledge. This suggests that the conventional Kullback-Leibler\,(KL) constraints between the policy model and the reference model\,\citep{ppo, dpo}, which is based solely on the tokens appearing in preferred or dispreferred responses\,\citep{smaug_dpop, ipo, scerets_rlhf_ppo}, fail to prevent the forgetting that occurs during personalized preference optimization.

To address this issue, we start by analyzing the initial responses, referred to as base responses, of instruction-tuned models to the given prompts and observe how their likelihood of producing these responses changes over training steps. We discover that personalizing preferences to enhance the distinction between preferred and dispreferred responses not only diminishes the likelihood of producing the dispreferred responses, but also lowers the likelihood of generating these base responses.

We hypothesize that aligning the reference and policy models, especially focusing on the tokens that appear in the base response, is essential for preserving global knowledge and ensuring general alignment. To this end, we introduce a novel preference optimization method named as Base-Anchored Preference Optimization\,(BAPO). BAPO aims to maintain the likelihood that the policy model will produce a base response originating from the reference model during personalized preference optimization. 

\noindent Our main contributions are summarized as follows:
\begin{itemize}
    \vspace{-3pt}
    \item We systematically analyze how diverse user preferences affect the global knowledge and alignment of instruction-tuned LLMs, finding that the extent of forgetting varies significantly with preference type. (\textbf{\Autoref{section:motivation}})
    \vspace{-3pt}
    \item We propose a novel preference optimization method, BAPO, which utilizes the base response from the reference model to preserve existing knowledge in instruction-tuned LLMs during personalized preference optimization. (\textbf{\Autoref{section:bapo}})
    \vspace{-2pt}
    \item We validate the efficacy of BAPO across various setups, demonstrating its effectiveness in preserving global knowledge and general alignment while adapting to diverse personalized preferences. (\textbf{\Autoref{section:experiment}})
    \vspace{-1pt}
\end{itemize}

\section{Personalized Preference Optimization}
\label{section:motivation}

In this section, we first introduce preference optimization for LLM alignment. Next, we examine how personalized preferences impact the existing knowledge of instruction-tuned LLMs. We suggest that the typical KL-constraint in preference optimization is not effective in preventing forgetting.

\subsection{Preliminary: Preference Optimization}
Consider a dataset of pairwise preferences, denoted as $\mathcal{D} = \{x^{i},\; y_w^i,\; y_l^i\}_{i=1}^{N}$. In this dataset, for each prompt $x^i$, the responses $y_w^i$ and $y_l^i$ represent the preferred\,(i.e., chosen) and dispreferred\,(i.e., rejected) responses, respectively. Our goal is to optimize the policy model $\pi_{\theta}(y|x)$ to maximize the expected value of the ideal reward function $r^*(x,\;y)$ that aligns with human preferences:
\begin{equation}
\pi^{*} = \underset{\pi_{\theta}}{\argmax}\;\mathbb{E}_{y \sim \pi_{\theta}(\,\cdot\,|x)}\left[r^{*}(x,y) \right]\,.
\end{equation}
A common approach to modeling the reward function is using the Bradley-Terry model\,\citep{bradley_terry}, which models the human preference distribution $p^{*} (y_1 \succ y_2 \;| \;x)$ as follows:
\begin{equation}
    \frac{\mathrm{exp}(r^{*}(x, y_w))}{\mathrm{exp}(r^{*}(x, y_w)) + \mathrm{exp}(r^{*}(x, y_l))}\,.
    \label{eq:bradley_terry}
\end{equation}
\noindent Note that the Bradley-Terry model assumes that for each prompt $x$, the paired comparison probabilities $p(y_w \succ y_l \;| \;x)$ reflect a consistent human preference ordering across all possible responses, depending solely on the reward difference between responses $r^*(x, y_w) - r^*(x, y_l)$.

\paragraph{RLHF} Using the reward function defined in \autoref{eq:bradley_terry},
Reinforced Learning from Human Feedback\,(RLHF)\,\citep{rlhf, learning_to_summarize_rlhf} initially trains a reward model $r_{\phi}(x, y)$ that produces a single scalar prediction for the reward value. In the subsequent RL phase, this reward model guides the LLM to align the learned preference with the reference model $\pi_{\text{ref}}$, which has undergone supervised fine-tuning (SFT) from a pre-trained LLM as follows:
\begin{align}
\mathcal{L}_{\text{RLHF}} = & -\mathbb{E}_{x \sim \mathcal{D}, y \sim \pi_{\theta}(y|x)} \left[r_{\phi}(x, y) \right. \notag \\
& \left. - \beta \mathbb{D}_{\text{KL}}\left[ \pi_{\theta}(y|x) \Vert \pi_{\text{ref}}(y|x) \right] \right].
\label{eq:rlhf}
\end{align}
\noindent where $\beta$ corresponds to the regularization strength of KL-Divergence between the policy model $\pi_{\theta}$ and the reference model $\pi_{\text{ref}}$.

\paragraph{DPO} By simplifying \autoref{eq:rlhf}, Direct Preference Optimization\,(DPO)\,\citep{dpo} optimizes the maximum likelihood of the policy model $\pi_{\theta}$ without the need to train a separate explicit reward model as follows:
\begin{align}
    \mathcal{L}_{\text{DPO}} = & -\mathbb{E}_{(x, y_{w}, y_{l}) \sim \mathcal{D}}  \bigg[ \log \sigma \left( \beta \log \frac{\pi_{\theta}(y_w|x)}{\pi_{\text{ref}}(y_w|x)} \right. \notag \\
    &\qquad \qquad \quad \left. - \beta \log \frac{\pi_{\theta}(y_l|x)}{\pi_{\text{ref}}(y_l|x)} \right) \bigg]\,.
\end{align}
\noindent Here, $\beta$ represents the KL-regularization strength in RLHF. Note that in both RLHF and DPO, this KL-constraint depends only on the tokens appearing in $y_w$ and $y_l$, the responses directly related to the preference ranking comparison.

\subsection{Forgetting from Personalization}

To understand how preference heterogeneity affects the extent of forgetting, we conduct an experimental study using heterogeneous preference datasets: P-Soups\,\citep{psoups} and DSP\,\citep{domain_specific_preference}. We fine-tune the instruction-tuned Phi-3-mini\,\citep{phi3} model using DPO\,\citep{dpo} with LoRA\,\citep{lora}. Please refer to the detailed setups provided in \Autoref{section:experiment}.
\begin{figure}[ht!]
    \centering
    \includegraphics[width=0.478\textwidth]{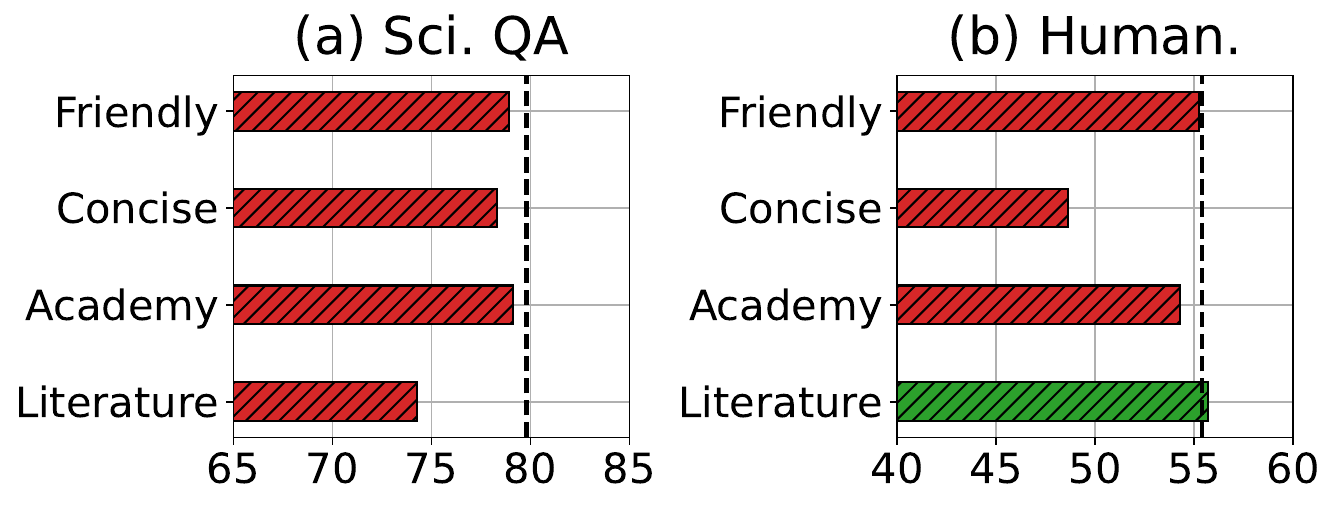}
    \vspace{-16pt}
    \caption{Performance on \textit{Global Knowledge}: (a) Science QA and (b) MMLU - Humanities after personalization on diverse preferences. The black vertical dotted line indicates the base model performance.}
    \label{fig:variation_global_knowledge}
\end{figure}
\begin{figure}[ht!]
    \vspace{-12pt}
    \centering
    \includegraphics[width=0.478\textwidth]{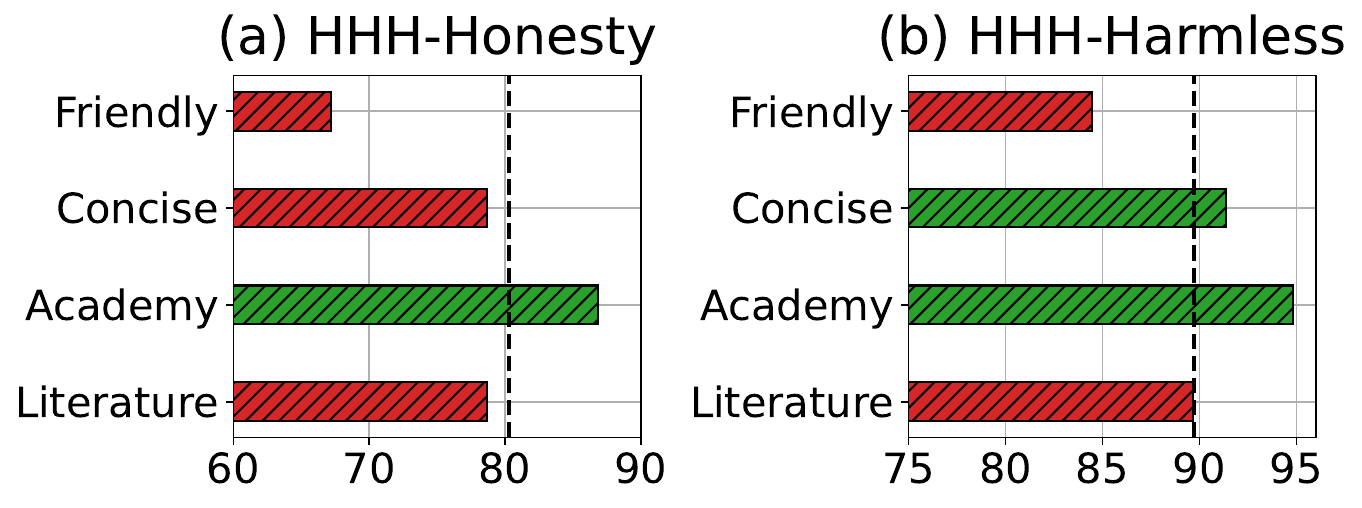}
    \vspace{-15pt}
    \caption{Performance on \textit{General Alignment}: (a) HHH-Honesty and (b) HHH-Harmless after personalization on diverse preferences. The black vertical dotted line indicates the base model performance.}
    \label{fig:variation_general_alignment}
    \vspace{-2pt}
\end{figure}

In \autoref{fig:variation_global_knowledge} and \autoref{fig:variation_general_alignment}, we evaluate performance changes after optimizing for specific preference types and present some representative results. The first two rows of each figure depict specific style preferences (e.g., Friendly or Concise) from the P-Soups dataset, while the last two rows showcase specific domain preferences (e.g., Academy or Literature) from the DSP dataset.

Our analysis reveals that the extent of forgetting global knowledge varies significantly with the prioritized preference type. For example, as shown in \autoref{fig:variation_global_knowledge}, personalizing for the \textit{Literature} domain preference significantly decreases performance on the Science QA\,\citep{science_qa} datasets, while the \textit{Academy} domain preference results in a smaller decline. Similarly, in the Humanities section of the MMLU datasets\,\citep{mmlu}, preferring the \textit{Friendly} style leads to a lesser decline in performance, whereas the \textit{Concise} style preference causes a substantial drop. This variation is not limited to global knowledge but also extends to general alignment. For example, in \autoref{fig:variation_general_alignment}, the \textit{Friendly} style preference significantly compromises Honesty in the HHH-Alignment\,\citep{hhh_alilgnment} datasets. On the other hand, favoring the \textit{Academy} domain preference rather improves it.

\subsection{Knowledge in Base Response}

The significant variation in performance after personalized preference optimization suggests that the typical KL-divergence constraints\,\citep{rlhf, dpo, scerets_rlhf_ppo} used in general preference optimization\,\citep{hh_rlhf, zephyr} still suffer from forgetting induced by preference heterogeneity.

We hypothesize that the original response from the initial reference model $\pi_{\text{ref}}$, which contains intact global knowledge and aligns with general alignment, is influenced by learning to meet diverse individual preferences. We take a closer look at personalized preference optimization to understand how adapting to heterogeneous preferences affects the likelihood of generating specific responses, represented by $\log [\pi_{\theta}(y_{(.)}|x) - \pi_{\text{ref}}(y_{(.)}|x)]$.

\begin{figure}[t!]
    \vspace{-9pt}
    \centering    \includegraphics[width=0.482\textwidth]{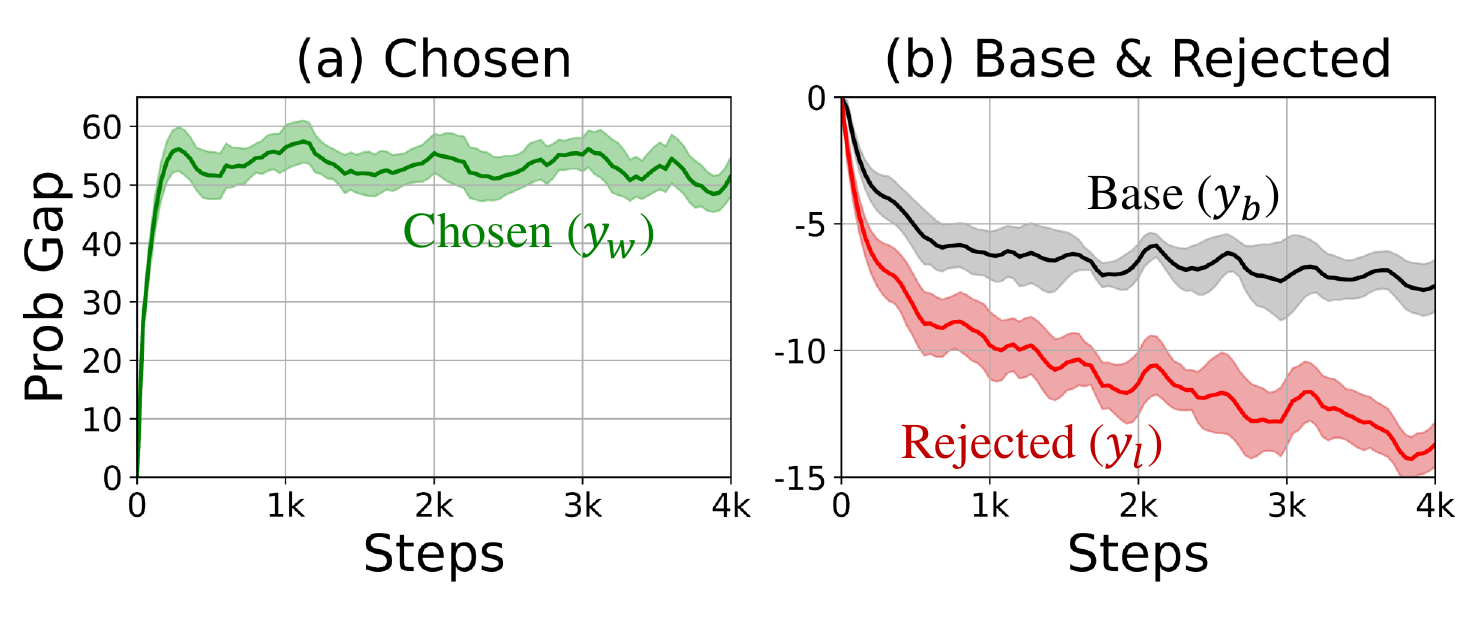}
    \vspace{-24pt}
    \caption{Average difference in reference model\,($\pi_{\text{ref}})$ and policy model\,($\pi_{\theta}$) log probabilities for \textbf{\textcolor{green!50!black}{Chosen}}, \textbf{Base}, and \textbf{\textcolor{red!80!black}{Rejected}} responses during personalization across four domain preferences in DSP datasets.}
    \vspace{-10pt}
    \label{fig:motive_base_response}
\end{figure}

The observations in \autoref{fig:motive_base_response} verify our conjecture. Personalizing preferences to enhance the distinction between the chosen response $y_w$ (i.e., preferred) and the rejected response $y_l$ (i.e., dispreferred) not only reduces the likelihood of producing the rejected response but also lowers the likelihood of generating base responses $y_b$. In this context, KL-divergence constraints between the reference and policy models on tokens found in these chosen and rejected responses do not help maintain the likelihood of base responses. Based on our findings, we consider leveraging the tokens appearing in the base responses to encourage knowledge preservation during personalized preference optimization. Please see more details in \Autoref{appendix:log_probs_responses}.
\begin{figure}[t!]
    \centering
    \includegraphics[width=0.45\textwidth]{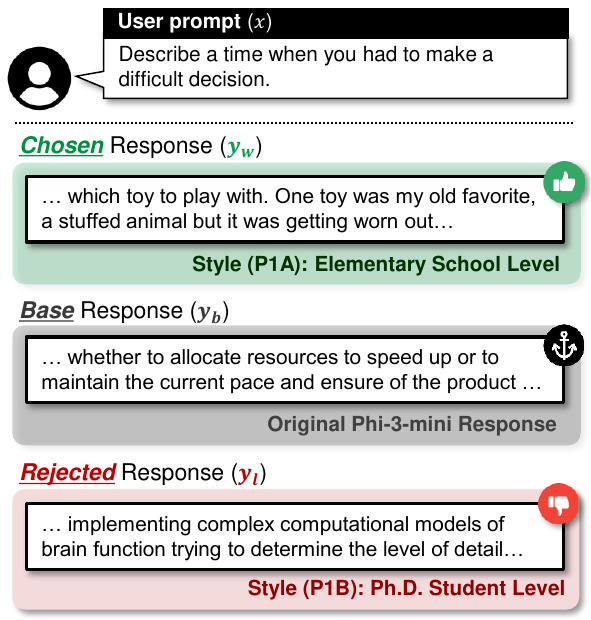}
    \vspace{-4pt}
    \caption{An example of responses for a user with P1A\,(elementary school level) style preference. The full example is provided in \autoref{appendix:base_full_example}}
    \label{fig:base_response_example}
\end{figure}

\section{Proposed Method: BAPO}
\label{section:bapo}

In this section, we introduce Base-Anchored Preference Optimization\,(BAPO). Our primary motivation is to use the initial response from the instruction-tuned model before it undergoes personalized preference optimization. By anchoring the policy model to this initial response, the personalized policy model can effectively retain its original global knowledge and general alignment while still accommodating diverse user preferences.

\subsection{Base-Anchored Preference Optimization}

Consider the base response $y_b$ from the reference model $\pi_\text{ref}$ and the policy model $\pi_{\theta}$, which we fine-tune for personalized preferences. The example in \autoref{fig:base_response_example} showcases how the chosen, base, and rejected responses differ for the same given user prompt. The core concept of BAPO is to preserve the knowledge contained in this base response during the optimization for diverse preferences. 

\paragraph{Base Anchor} BAPO ensures that the policy model's likelihood of producing the base response $y_b$ ($\pi_{\theta}(y_b|x)$) remains closely aligned with that of the reference model ($\pi_\text{ref}(y_b|x)$):
\begin{equation}
    \mathcal{L}_{\text{Anchor}} = \max \left(0,\;  \log\frac{{\pi}_{\text{ref}}(y_b|x)}{{\pi}_{\theta}\,(y_b|x)} \right)\,.
\end{equation}
\noindent Note that the base anchor loss $\mathcal{L}_{\text{Anchor}}$ becomes 0 if the policy model $\pi_{\theta}$ assigns a higher likelihood to the base response $y_b$ than the reference model does ($\pi_{\theta}(y_b|x) > \pi_{\text{ref}}(y_b|x)$). Intuitively, if the policy model is already more confident in the base response than the reference model, there's no need to penalize it further. The BAPO objective $\mathcal{L}_{\text{BAPO}}$ is defined as follows:
\begin{align}
    \vspace{-2pt}
    \mathcal{L}_{\text{BAPO}} = \mathcal{L}_{\text{DPO}} + {\lambda}\cdot \mathcal{L}_{\text{Anchor}}\,.
    \vspace{-2pt}
\end{align}
Here, $\lambda$ controls the strength of the anchoring effect. In our main experiments, we set $\lambda$ to 5.

\subsection{Theoretical Analysis}
We assess the impact of BAPO on personalized preferences by analyzing how information from base responses aids in aligning personal preferences. We assume linear utility and reward functions, with their respective unknown parameters having distinct nonzero components.

\vspace{3pt}
\begin{assumption}
\label{assumption1:existence_of_utilities}
A utility function $G^{\star}$ exists for general alignment with global knowledge, and a reward function $L^{\star}$ measures personal alignment.
\end{assumption}

\vspace{1pt}
\begin{assumption}
\label{assumption2:linear_model_independence}
For the response $y$ and context $x$, the functions $G^{\star}$ and $L^{\star}$ are linear, defined as: $G^{\star}(x,y) = \langle \boldsymbol{\phi}(x,y),\boldsymbol{\theta}_{\star}^{G}\rangle$ and $L^{\star}(x,y) = \langle \boldsymbol{\phi}(x,y),\boldsymbol{\theta}_{\star}^{L}\rangle$ where $\boldsymbol{\phi}(x,y)$ is a $d$ dimensional feature vector of $y$ and $x$. Additionally, $\boldsymbol{\theta}_{\star}^{G},\boldsymbol{\theta}_{\star}^{L}$ have non-intersecting nonzero components on d-k and the k dimensions respectively.
\end{assumption}
\vspace{2pt}

\begin{proposition}
\label{prop:sample_complexity}
Given the information of ${\theta}^{}_{G}$ is known. Then, the sample complexity for estimating ${\theta}^{}_{L}$ reduces from $O(\sqrt{d})$ to $O(\sqrt{k})$.
\end{proposition}
\vspace{4pt}

The proof is provided in \autoref{appendix:proof}. This proposition implies that the $k$-dimensional subspace, which governs personalized rewards, is typically much smaller than the $(d-k)$-dimensional subspace responsible for general alignment. Consequently, this reduction in the complexity of the parameter space influencing personalized rewards allows for more efficient estimation with fewer samples. In practice, personalization is often driven by a smaller, critical set of features compared to those affecting broader alignment criteria.

\begin{figure}[h!]
    \centering
    \includegraphics[width=0.482\textwidth]{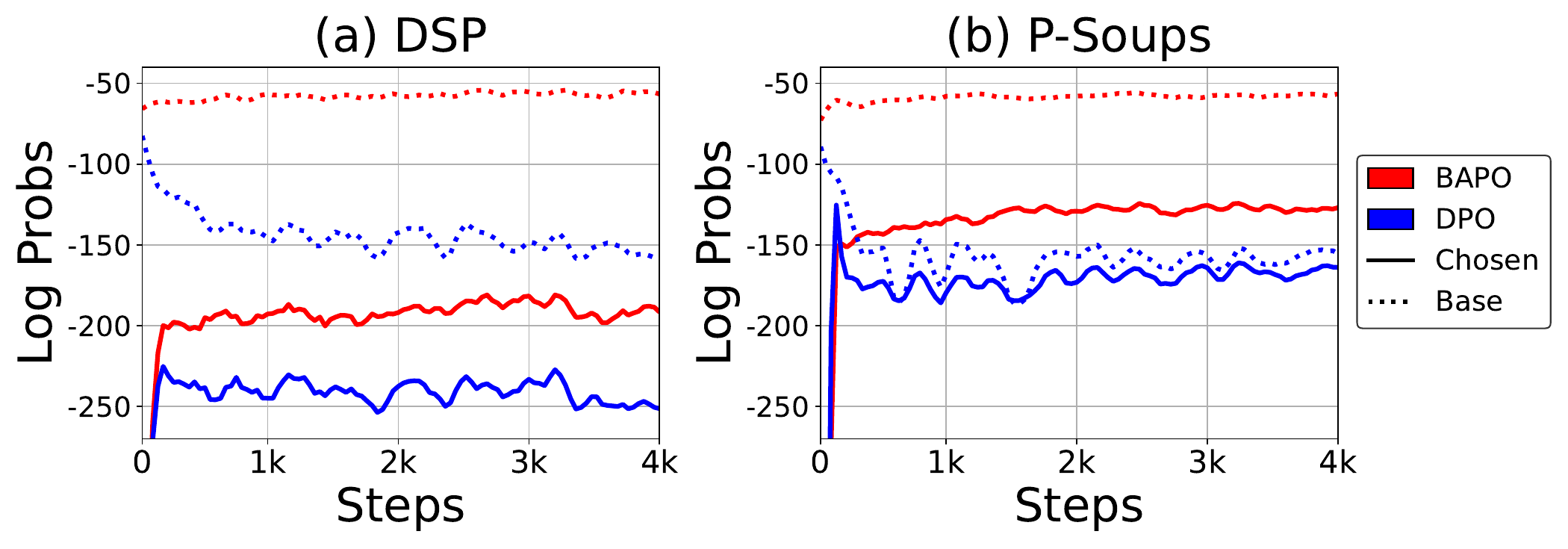}
    \vspace{-16pt}
    \caption{Evolution of log probabilities for \textit{Chosen} and \textit{Base} responses during preference optimization. The results are averaged across different preference types.}
    \label{fig:logps_chosen_base}
    \vspace{-12pt}
\end{figure}

More intuitively, the theoretical analysis in \textcolor{red}{Proposition}\,\autoref{prop:sample_complexity} suggests that identifying the parameter dimensions crucial for global knowledge and general alignment (${\theta}_{G}$) can enhance the process of optimizing personalized preferences. This is achieved by focusing on the parameter dimensions essential for personalization (${\theta}_{L}$), under the assumption that ${\theta}_{G}$ and ${\theta}_{L}$ are separable. The results in \autoref{fig:logps_chosen_base} empirically support our claim. We analyze the changes in log probabilities of responses under DPO and BAPO. The experimental details are provided in \Autoref{section:experiment}. While the log probabilities for the \textit{base} response decrease under DPO, BAPO maintains them, enhancing stability throughout the preference optimization process. This consistency enables the model to assign significantly higher log probabilities to the \textit{chosen} response, thus accelerating the learning process. 

\section{Experiment}
\label{section:experiment}
\vspace{-2pt}
\subsection{Experimental Setups}

\paragraph{Datasets} We use two preference datasets for personalized preference optimization: P-Soups\,\citep{psoups} and DSP\,\citep{domain_specific_preference}.
\vspace{-2pt}
\begin{itemize}
    \item \textbf{Personalized Soups (P-Soups)} include \textit{Style} preferences , organized into three dimensions: P1, P2, and P3. Each dimension features two contrasting types, A and B. In \autoref{tab:psoups_preference}, we briefly describe the preference types.
    
    \vspace{-5pt}
    \item \textbf{Domain Specific Preference\,(DSP)} includes \textit{Domains} preferences: Academy, \textit{Business}, Entertainment, and Literature\,\&\,Art.
    \vspace{-3pt}
\end{itemize}
Each dataset is composed of user queries and a set of responses for each query. In our pairwise preference format, for each user query\,$x$, we select a response that aligns with a specific preference as the chosen response\,$y_w$. Responses from other preferences are designated as rejected responses\,$y_l$. More details are provided in \Autoref{appendix:datasets}.

\begingroup
\setlength{\tabcolsep}{2.0pt} 
\renewcommand{\arraystretch}{1.125}
\begin{table}[ht!]
\caption{Response Preferences of the P-Soups dataset.}
\vspace{-2pt}
\label{tab:psoups_preference}
\centering
\small
\begin{tabular}{lcl}
\toprule
\multicolumn{1}{c}{\textbf{Dimension}} & \textbf{Type} & \multicolumn{1}{c}{\textbf{Response Preference}}          \\ 
\hline\hline
\multirow{2}{*}{(P1) Expertise}        & A             & Elementary school level.\\
                                       & B             & PhD-level expertise in the field. \\ 
\hline
\multirow{2}{*}{(P2) Verbosity}        & A             & Concise, without being verbose.                \\
                                       & B             & Informative and fully detailed.\\ 
\hline
\multirow{2}{*}{(P3) Style}            & A             & Friendly, witty and humorous.     \\
                                       & B             & Answer in an unfriendly manner.                                \\
\bottomrule
\end{tabular}
\end{table}
\endgroup
\vspace{-8pt}

\begingroup
\setlength{\tabcolsep}{5.3pt} 
\renewcommand{\arraystretch}{1.17}
\begin{table*}[ht!]
\caption{Performance on global knowledge datasets after fine-tuning for personalized preferences. The term 'Base' refers to the initial performance of the Phi-3-mini model before personalization. Values in parentheses represent the standard deviation across different preference types: 6 for the P-Soups datasets and 4 for the DSP datasets.}
\vspace{-5pt}
\label{tab:main_global_knowledge}
\small
\centering
\begin{tabular}{llllllllll} 
\bottomrule
\multicolumn{10}{c}{\textbf{Preference Dataset: P-Soups\,\citep{psoups}}} \\ 
\hline
\multicolumn{1}{c}{\multirow{2}{*}{\textbf{Method}}}
& \multirow{2}{*}{\textbf{PIQA}}
& \multirow{2}{*}{\textbf{SIQA}}  
& \multirow{2}{*}{\textbf{ARC-c}}  
& \multirow{2}{*}{\textbf{Sci.\,QA}} 
& \multirow{2}{*}{\textbf{Comm.}} 
& \multicolumn{4}{c}{\textbf{MMLU}} \\
\multicolumn{1}{c}{} & & & &  & & STEM & Social & Human & Other \\
\hline\hline
Base & 78.0  & 72.0  & 81.9 & 79.8 &  46.0 & 47.7 & 62.7 & 55.4 & 65.3 \\ 
\hline
DPO  & 77.0${}_{(2.8)}$  & 69.7${}_{(0.1)}$  & 81.6${}_{(0.9)}$ & 78.5${}_{(2.1)}$ &  68.7${}_{(1.6)}$ & 49.2${}_{(1.9)}$ & 65.7${}_{(8.4)}$ & 53.7${}_{(2.5)}$ & 64.8${}_{(5.5)}$ \\
RSO  & 76.4${}_{(2.4)}$  & 71.1${}_{(0.7)}$  & 82.3${}_{(1.3)}$ & 80.8${}_{(1.5)}$ &  70.5${}_{(0.7)}$ & 50.3${}_{(1.2)}$ & 67.6${}_{(6.5)}$ & \textbf{55.3}${}_{(0.5)}$ & 66.9${}_{(1.9)}$ \\
IPO  & 62.9${}_{(11.0)}$  & 50.2${}_{(13.6)}$  & 51.2${}_{(24.2)}$ & 52.2${}_{(20.8)}$ &  41.8${}_{(19.2)}$ & 38.2${}_{(11.3)}$ & 51.4${}_{(19.8)}$ & 39.8${}_{(10.3)}$ & 48.2${}_{(16.6)}$ \\
DPOP & 76.1${}_{(1.7)}$  & 70.8${}_{(0.8)}$  & 81.8${}_{(1.7)}$ & 80.6${}_{(1.1)}$ & 70.6${}_{(1.4)}$ & 50.5${}_{(1.6)}$ & 68.3${}_{(4.5)}$ & 54.7${}_{(0.8)}$ & \textbf{67.4}${}_{(1.4)}$ \\
ORPO & 66.6${}_{(6.0)}$  & 61.5${}_{(4.0)}$  & 71.5${}_{(2.9)}$ & 68.5${}_{(6.4)}$ &  60.7${}_{(2.9)}$ & \textbf{51.2${}_{(1.5)}$} & \textbf{70.0${}_{(2.7)}$} & 50.5${}_{(0.8)}$ & 65.1${}_{(1.1)}$ \\
\hline
\textbf{BAPO} 
& \textbf{78.0${}_{(1.5)}$}
& \textbf{71.6${}_{(0.8)}$}
& \textbf{82.5${}_{(1.1)}$}
& \textbf{81.0${}_{(0.6)}$}
& \textbf{71.5${}_{(0.7)}$} 
& 49.7${}_{(1.0)}$ 
& 66.1${}_{(3.0)}$ 
& 55.1${}_{(0.6)}$ 
& 66.4${}_{(1.0)}$ \\
\bottomrule
\multicolumn{10}{c}{\textbf{Preference Dataset: DSP\,\citep{domain_specific_preference}}}\\
\hline
\multicolumn{1}{c}{\multirow{2}{*}{\textbf{Method}}}
& \multirow{2}{*}{\textbf{PIQA}}
& \multirow{2}{*}{\textbf{SIQA}}    
& \multirow{2}{*}{\textbf{ARC-c}}  
& \multirow{2}{*}{\textbf{Sci.\,QA}} 
& \multirow{2}{*}{\textbf{Comm.}} 
& \multicolumn{4}{c}{\textbf{MMLU}} \\
\multicolumn{1}{c}{} & & & & & & STEM & Social & Human & Other \\
\hline\hline
Base & 78.0  & 72.0  & 81.9 & 79.8 & 46.0 & 47.7 & 62.7 & 55.4 & 65.3 \\ 
\hline
DPO  & 77.6${}_{(0.6)}$  & 70.2${}_{(1.3)}$  & 81.7${}_{(1.1)}$ & 78.5${}_{(3.1)}$ & 69.1${}_{(0.9)}$ & 49.9${}_{(3.0)}$ & 65.2${}_{(9.2)}$ & 54.1${}_{(1.9)}$ & 66.3${}_{(2.2)}$ \\
RSO  & 77.2${}_{(0.9)}$  & 71.2${}_{(0.5)}$  & 81.9${}_{(1.0)}$ & 80.2${}_{(0.9)}$ &  70.4${}_{(1.1)}$ & 50.5${}_{(2.0)}$ & 66.8${}_{(6.2)}$ & \textbf{55.9${}_{(0.6)}$} & 66.7${}_{(1.5)}$ \\
IPO  & 50.9${}_{(1.3)}$  & 35.5${}_{(3.1)}$  & 19.8${}_{(13.1)}$ & 24.8${}_{(9.4)}$ & 25.8${}_{(6.6)}$ & 28.0${}_{(1.9)}$ & 30.2${}_{(4.8)}$ & 30.3${}_{(2.3)}$ & 30.0${}_{(2.5)}$ \\
DPOP & 77.9${}_{(1.5)}$  & 70.9${}_{(1.0)}$  & 81.6${}_{(2.4)}$ & 80.1${}_{(0.4)}$ &  69.8${}_{(1.5)}$ & 50.4${}_{(1.9)}$ & 67.4${}_{(7.4)}$ & 55.1${}_{(1.4)}$ & 67.2${}_{(1.5)}$ \\
ORPO & 66.6${}_{(6.0)}$  & 61.5${}_{(4.0)}$  & 71.5${}_{(2.9)}$ & 68.9${}_{(6.4)}$ & 60.7${}_{(2.9)}$ & \textbf{51.1${}_{(1.5)}$} & \textbf{69.9${}_{(2.7)}$} & 50.5${}_{(0.7)}$ & 65.1${}_{(1.1)}$ \\
\hline
\textbf{BAPO} 
& \textbf{78.0${}_{(1.2)}$}
& \textbf{71.7${}_{(0.3)}$}  
& \textbf{83.2${}_{(0.5)}$} 
& \textbf{80.9${}_{(1.1)}$} 
& \textbf{71.1${}_{(0.7)}$} 
& 50.2${}_{(0.9)}$ 
& 66.1${}_{(2.9)}$ 
& 55.6${}_{(0.4)}$ 
& \textbf{67.1${}_{(0.4)}$} \\
\bottomrule
\end{tabular}
\vspace{-8pt}
\end{table*}
\endgroup

\paragraph{Learning Setups} In our main experiments, we primarily use a Phi-3 model\,\citep{phi3}, specifically its instruction-tuned version referred to as \textit{Phi-3-mini-128k-instruct}, with 3.82 billion parameters. This model has been enhanced with DPO to align with general human preferences and safety guidelines, following the SFT stage. 
Each personalized model, aimed at aligning with a specific preference type, is fine-tuned using Q-LoRA\,\citep{qlora}, a quantized variant of LoRA\,\citep{lora}. We utilize a 4-bit normalized float\,(nf4) and double quantization with bf-16 to enhance computational efficiency. The LoRA settings, including a rank of $r$\,=\,32 and $\alpha$\,=\,64 with a dropout rate of 0.05, are applied to all linear layer weights of the model. Each personalized model is trained for a single epoch on feedback pairs using an effective batch size of 8. The learning rate, set at 5e-5, follows the conventional training recipe\,\citep{alignment_handbook, zephyr}. The learning rate is decayed using a cosine scheduler\,\citep{cosine}.

\paragraph{Evaluation} To evaluate the extent of forgetting after personalized preferences optimization, we divide the datasets into two categories: (i) Global Knowledge and (ii) General Alignment. In Global Knowledge, we assess the model's prior knowledge of world understanding through closed-book question-answering tasks. More specifically, we use commonsense datasets such as PIQA\,\citep{piqa}, SIQA\,\citep{social_i_qa}, Arc-Challenge\,\citep{arc}, Science QA\,\citep{science_qa}, Commonsense QA\,\citep{commonsense_qa}, and 5-shot MMLU\,\cite{mmlu}. Since the Science QA dataset includes visual tasks, we utilize text-only questions from this dataset. For assessing General Alignment, we conduct evaluations using the HHH-Alignment\,\citep{hhh_alilgnment} datasets, which consist of categories focused on helpfulness, harmlessness, and honesty.

\subsection{Performance on Knowledge Preservation}

In \autoref{tab:main_global_knowledge}, we evaluate how fine-tuning the Phi-3-mini model affects its performance in the \textit{Global Knowledge} across diverse preferences. This evaluation includes the DPO\,\citep{dpo} and other preference optimization methods such as RSO\,\citep{rso}, IPO\,\citep{ipo}, DPOP\,\citep{smaug_dpop}, and ORPO\,\citep{orpo}. The 'Base' in the table indicates the initial performance of the Phi-3-mini model.

The results show that the baseline methods significantly declines the performance on global knowledge datasets, while our BAPO method effectively maintains consistent performance across various evaluated datasets. We highlight that BAPO has advantageous characteristics, keeping performance variations due to preference heterogeneity remarkably low. In contrast to the baseline method, which exhibits high fluctuation and large variance across varying preferences, BAPO provides more reliable results with minimal variation.

We observe that personalized preferences do not necessarily lead to forgetting. In fact, they can sometimes enhance performance on certain datasets based on the types of preferences involved. For example, fine-tuning with domain-specific preferences in DSP datasets often results in improved performance on MMLU datasets. Notably, the ORPO method, which does not use a reference model during preference optimization, consistently outperforms other approaches in such cases. This implies there exists a natural trade-off in the use of reference models between preserving existing knowledge and acquiring new knowledge.

\subsection{Performance on Alignment}

\paragraph{General Alignment} \autoref{tab:main_alignment} presents the evaluation of general and personalized alignment after the personalized preference optimization. Ideally, the personalized model should maintain general alignment while accommodating its specific personalized preferences. We first evaluate whether the fine-tuned models maintain general alignment after personalization by using the HHH-Alignment datasets \citep{hhh_alilgnment}. The results show that BAPO effectively preserves the level of general alignment seen in the initial reference model, with minimal variation across diverse preference types.

\paragraph{Personalized Alignment} We evaluate personalized alignment by measuring Reward Accuracy, which assesses whether the policy model prefers the selected response $y_w$ over the dis-preferred response $y_l$ for user prompts $x$ in the validation set. The results show that BAPO more effectively accommodate personalized performance compared to other baselines. This demonstrates that utilizing base responses can boost sample efficiency for reward signals related to personalized preferences.

\begingroup
\setlength{\tabcolsep}{4.0pt} 
\renewcommand{\arraystretch}{1.15}
\begin{table}[ht!]
\caption{Performance on general/personalized alignment after fine-tuning for personalized preferences.}
\vspace{-5pt}
\label{tab:main_alignment}
\small
\centering
\begin{tabular}{lcccc} 
\toprule
\multicolumn{5}{c}{\textbf{Preference Datasets: P-Soups}}                                                                                                         \\ 
\hline
\multicolumn{1}{c}{\multirow{2}{*}{\textbf{Method }}} & \multicolumn{3}{c}{\textbf{HHH-Alignment }}                        & \multirow{2}{*}{\textbf{Rwd Acc. }}  \\
\multicolumn{1}{c}{}                                  & Helpful              & Harmless             & Honest               &                                      \\ 
\hline\hline
Base  & 84.7 & 89.7 & 80.3 & -  \\ 
\hline
DPO                                                   
& 81.4${}_{(6.3)}$  & 87.9${}_{(5.3)}$  & 76.0${}_{(4.8)}$                     
& 93.2${}_{(3.1)}$                                      
\\
RSO
& 83.1${}_{(3.6)}$  & \textbf{89.7}${}_{(3.6)}$  & 78.4${}_{(3.5)}$ 
& 93.0${}_{(3.1)}$  \\
IPO                                                   
& 70.3${}_{(11.7)}$ & 73.3${}_{(16.5)}$  & 68.0${}_{(14.2)}$                      
& 89.6${}_{(7.6)}$                                      
\\
DPOP                                                  
& 83.1${}_{(4.7)}$  & \textbf{89.7}${}_{(4.4)}$  & 79.2${}_{(2.9)}$                     
& 92.2${}_{(3.4)}$                                       
\\
ORPO                                                  
& 81.9${}_{(2.8)}$  & 83.6${}_{(5.5)}$  & 74.9${}_{(3.1)}$                    
& 84.6${}_{(2.1)}$                                    
\\
\hline
BAPO                                                  
& \textbf{84.0}${}_{(1.8)}$  & 87.9${}_{(4.2)}$  & \textbf{80.6}${}_{(1.2)}$                      
& \textbf{97.8}${}_{(2.3)}$                                      
\\ 
\bottomrule
\multicolumn{5}{c}{\textbf{Preference Datasets: DSP}}                                                                                                             \\ 
\hline
\multicolumn{1}{c}{\multirow{2}{*}{\textbf{Method }}} & \multicolumn{3}{c}{\textbf{HHH-Alignment }}                        & \multirow{2}{*}{\textbf{Rwd Acc. }}  \\
\multicolumn{1}{c}{}                                  & Helpful              & Harmless             & Honest               &                                      \\
\hline\hline
Base  & 84.7 & 89.7 & 80.3 & - \\ 
\hline
DPO                                                   
& 82.6${}_{(2.9)}$  & 91.8${}_{(3.6)}$  & 80.7${}_{(4.9)}$ 
& 87.1${}_{(4.0)}$ \\
RSO                                                   
& 81.6${}_{(1.3)}$  & 91.4${}_{(2.7)}$  & \textbf{81.4}${}_{(2.4)}$ 
& 87.1${}_{(4.3)}$ \\
IPO                                                   
& 72.9${}_{(7.3)}$  & 74.6${}_{(8.6)}$  & 68.0${}_{(7.6)}$
& 86.2${}_{(3.7)}$ \\
DPOP                                                  
& 72.9${}_{(7.3)}$  & 74.6${}_{(8.6)}$  & 68.0${}_{(7.6)}$
& 87.3${}_{(4.1)}$ \\ 
ORPO                                                  
& 83.1${}_{(1.4)}$  & 92.2${}_{(3.3)}$  & 77.1${}_{(3.8)}$ 
& 79.2${}_{(3.2)}$\\
\hline
BAPO                                                  
& \textbf{85.2}${}_{(1.6)}$  & \textbf{93.1}${}_{(1.4)}$  & 81.2${}_{(3.4)}$
& \textbf{97.0}${}_{(0.1)}$ \\
\bottomrule
\end{tabular}
\vspace{-12pt}
\end{table}
\endgroup

\subsection{Ablation Study}

\paragraph{Model Architecture} We conduct further experiments with the instruction-tuned Gemma-2B model\,\citep{gemma}, referred to as \textit{Gemma-2B-it}, which has also undergone RLHF for general alignment after the SFT stage. The results presented in \autoref{tab:ablation_gemma} validates the robust efficacy of BAPO across different model architectures.
\begingroup
\setlength{\tabcolsep}{2.0pt} 
\renewcommand{\arraystretch}{1.16}
\begin{table}[ht!]
\vspace{-1pt}
\caption{Performance of the Gemma-2B-it model after fine-tuning for personalized preferences. Scores for MMLU and HHH are averaged across all categories.}
\label{tab:ablation_gemma}
\vspace{-4pt}
\small
\begin{tabular}{lccccc}
\toprule
\multicolumn{6}{c}{\textbf{Preference Datasets: P-Soups}}\\
\hline
\textbf{Method} & 
\textbf{Sci.\,QA} & 
\textbf{Comm.} & 
\textbf{MMLU} & 
\textbf{HHH} & 
\textbf{Rwd Acc} \\ 
\hline\hline
Base   
& 51.9 
& 46.0
& 28.8
& 67.4
& -
\\
\hline
DPO\scriptsize{\,(P-Soups)}    
& 51.3
& 44.6
& 28.0
& 63.9
& \textbf{95.4}${}_{(3.5)}$             
\\
BAPO\scriptsize{\,(P-Soups)}
& \textbf{51.5}
& \textbf{44.9}
& \textbf{28.1}
& \textbf{64.2} 
& \textbf{95.4}${}_{(3.4)}$            
\\
\hline
DPO\scriptsize{\,(DSP)}
& 51.5 
& 45.4
& 28.3
& 66.1   
& 98.2${}_{(0.6)}$             
\\
BAPO\scriptsize{\,(DSP)}
& \textbf{51.6}
& \textbf{45.7}
& \textbf{28.5}
& \textbf{66.1}
& \textbf{98.3}${}_{(0.8)}$
\\
\bottomrule
\end{tabular}
\vspace{-12pt}
\end{table}
\endgroup

\begin{figure}[ht!]
    \vspace{-2pt}
    \centering
    \includegraphics[width=0.482\textwidth]{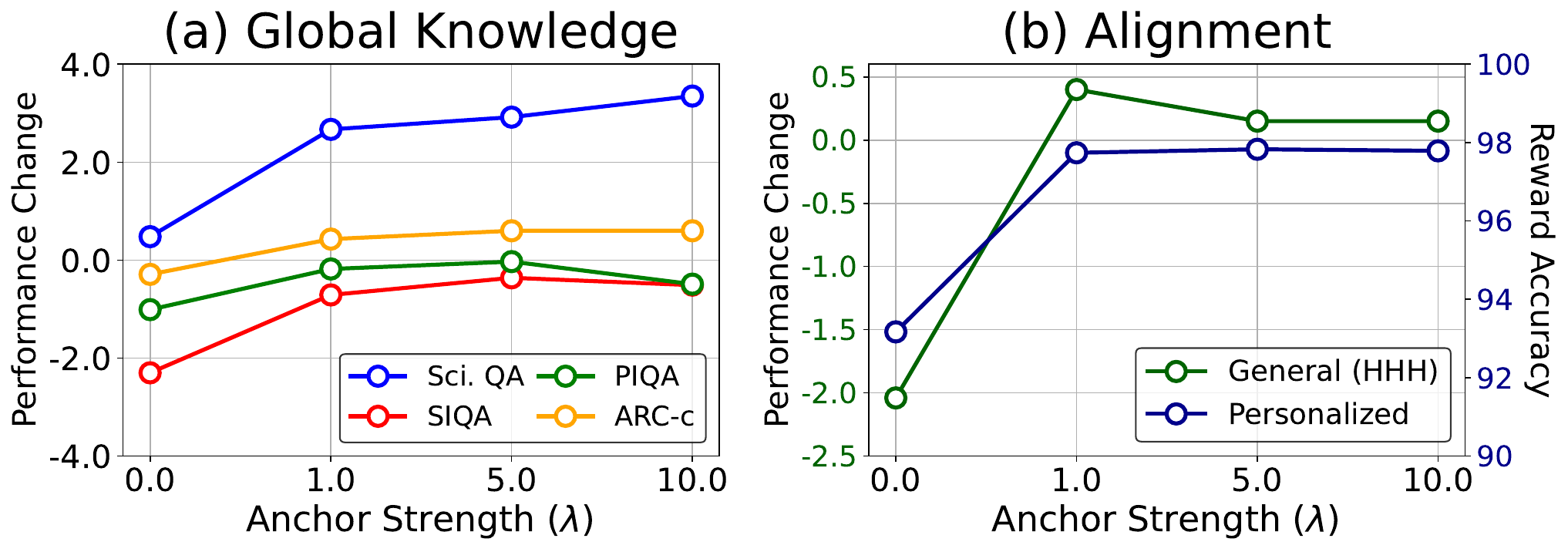}
    \vspace{-16pt}
    \caption{Performance of BAPO on P-Soups datasets with varying anchoring strength values $\lambda$. Note that setting $\lambda$\,=\,0 is equivalent to using the vanilla DPO.}
    \label{fig:ablation_bapo_lambda}
    \vspace{-8pt}
\end{figure}

\paragraph{Effect of Anchoring Strength} The impact of anchoring strength $\lambda$ on BAPO is illustrated in \autoref{fig:ablation_bapo_lambda}. The performance change is measured against the \textit{Base} model. As depicted in \autoref{fig:ablation_bapo_lambda}\textcolor{red}{(a)}, increasing anchoring strength generally enhances global knowledge preservation, although the effect plateaus at higher strengths. Additionally, \autoref{fig:ablation_bapo_lambda}\textcolor{red}{(b)} shows the results on alignment, indicating that while base anchoring in BAPO improves both types of alignment, its effectiveness also reaches a limit at higher strengths.

\begin{figure}[ht!]
    \centering
    \includegraphics[width=0.48\textwidth]{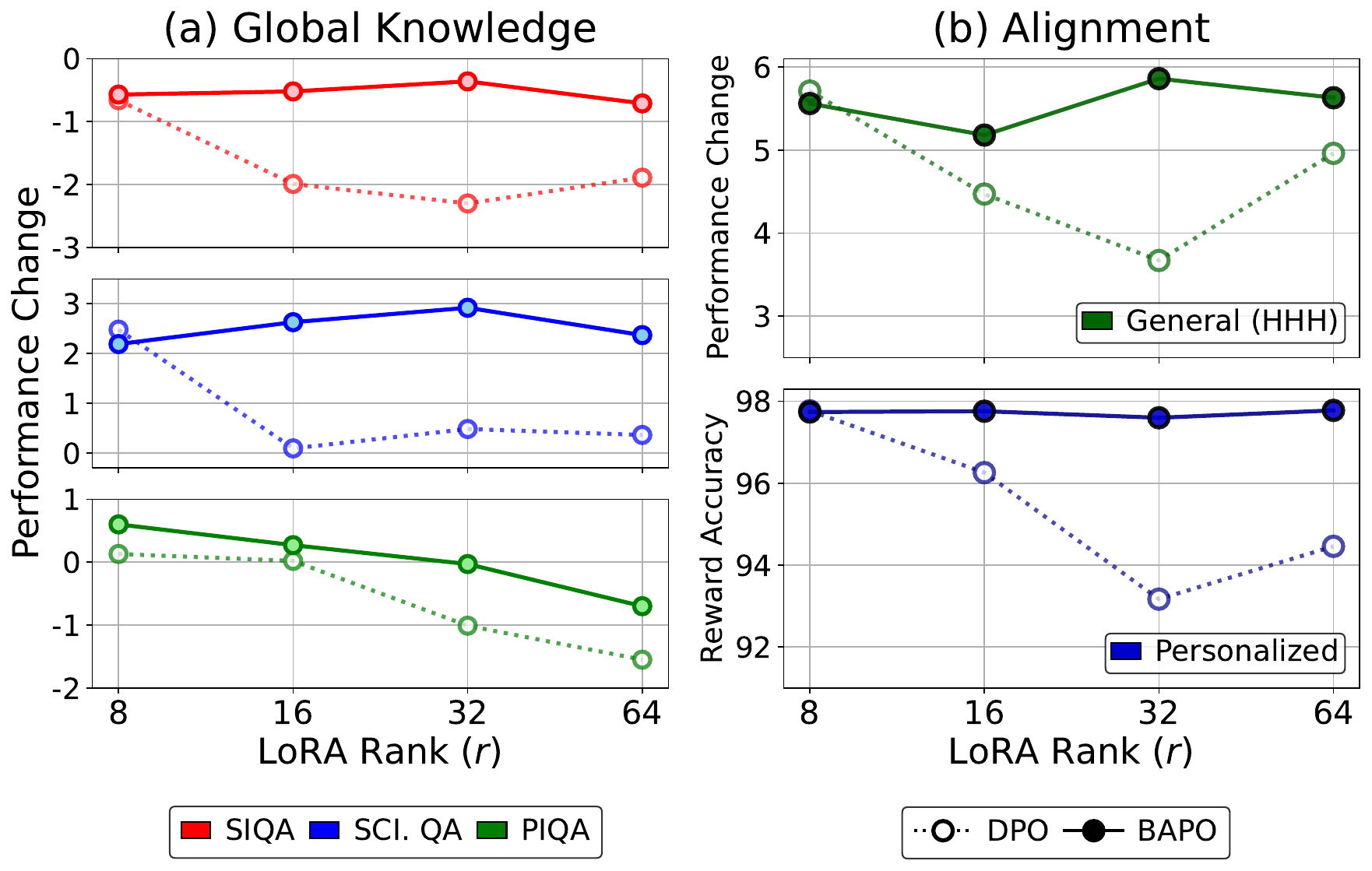}
    \vspace{-12pt}
    \caption{Performance of BAPO on P-Soups datasets with varying LoRA rank values $r$.}
    \label{fig:ablation_lora_rank}
    \vspace{-4pt}
\end{figure}

\paragraph{Effect of LoRA Rank} In \autoref{fig:ablation_lora_rank}, we explore the impact of varying LoRA rank $r$ on knowledge preservation and alignment, setting the scaling factor $\alpha$ such that $\frac{\alpha}{r} = 2$. As shown in \autoref{fig:ablation_lora_rank}\textcolor{red}{(a)} and \autoref{fig:ablation_lora_rank}\textcolor{red}{(b)}, an increase in rank $r$ leads to more pronounced forgetting in both global knowledge and general alignment. This finding supports recent research suggesting that a lower LoRA rank reduces the rate of learning but also minimizes forgetting\,\citep{lora_forgets_less}. Nevertheless, BAPO effectively preserves knowledge even as LoRA rank increases and enhances accommodation of personalized preferences at higher ranks, where the increased learning capacity could otherwise detract from learning personalized preferences.

\section{Discussion} 

We demonstrate that BAPO benefits from the separation between general alignment and personalized preferences. However, preserving the likelihood of the base response may sometimes hinder alignment with personalized preferences. For example, if a user prefers humorous but potentially harmful responses, and the base response is harmless, maintaining the base likelihood can conflict with achieving the user's preferred style. Thus, while we distinguish between 'Global Knowledge \& General Alignment' and 'Personalized Alignment,' their interaction is influenced by both the model's existing knowledge and the new, targeted personalization.

Our experiments show that this separation is advantageous for BAPO, as personalized preferences—whether in specific response styles (P Soups Datasets) or specific domains (DSP Datasets)—tend not to overlap significantly with the model’s general knowledge. Nevertheless, the challenge arises when preserving the base response's likelihood ($y_b$) compromises the ability to cater to personalized preferences. This situation parallels the Stability-Plasticity Dilemma in conventional Continual Learning\,\citep{stability1,stability2}, where the relationship between old and new knowledge influences the potential for forgetting.
\section{Related Work}
\vspace{-1pt}

\paragraph{Learning from Human Feedback} Often employed as the final stage of instruction tuning \citep{zephyr, mixtral, llama3}, learning from human feedback refines the policy language model to generate responses that align with human preferences\,\citep{alignment_survey, scerets_rlhf_ppo}. This process usually involves collecting human preferences for pairs of candidate responses to differentiate between those that are preferred and those that are dispreferred. The two main approaches used are Reinforcement Learning from Human Feedback\,(RLHF)\,\citep{learning_to_summarize_rlhf, rlhf} and Direct Preference Optimization\,(DPO)\,\citep{dpo}. RLHF involves training a separate reward model that is then utilized in the subsequent reinforcement learning phase\,\citep{hh_rlhf, zephyr}. In contrast, DPO does not rely on a reward model but directly establishes a mapping between the reward function and the optimization objective\,\citep{smaug_dpop, rso}. Our work specifically addresses the challenge of accommodating personalized preferences within the DPO framework. We focus on maintaining the likelihood of base responses from the reference model during the personalization process, ensuring the preservation of global knowledge and general alignment.

\paragraph{Forgetting in LLM Fine-tuning} The issue of forgetting previously acquired knowledge during the fine-tuning on heterogeneous data has been extensively discussed\,\citep{forgetting_survey, cl_survey}, highlighting a fundamental trade-off between preserving old knowledge and acquiring new knowledge\,\citep{cl1, cl2}. In the context of LLMs, recent research has shown that the knowledge from pre-training can be compromised by supervised fine-tuning\,(SFT) on instruction data\,\citep{loramoe, llm_sft_data_composition}. A similar issue, known as \textit{alignment tax}, occurs in preference optimization\,\citep{mitigating_alignment_tax, online_merging_rlhf}. While prior research has addressed the forgetting induced by the LLM fine-tuning\,\citep{lora_forgets_less, empirical_continual_finetuning}, the effects of accommodating specific user preferences and the impact of their heterogeneity remain largely unexplored. In our study, we investigate how personalized preference optimization affects both global knowledge and general alignment.

\paragraph{Personalized Preference in LLM} The use of a single scalar reward to represent user preferences presents a significant limitation when users have diverse and conflicting preferences\,\citep{alignment_survey, scerets_rlhf_ppo}. To address this issue, some studies have explored clustering users who presumably share the same reward\,\citep{maxmin_rlhf, personalization_preference_aggregation}. Others have considered defining a reward function with multiple objective dimensions\,\citep{morl_prompt, rewards_in_context} to achieve Pareto optimality among them\,\citep{cpo_controllable, panacea, dpa_morl}. Additionally, some approaches involve merging model parameters trained for each dimension to accommodate the diverse combinations expressed by those dimension\,\citep{psoups, reward_soups}. In our study, we focus on the impact of preference heterogeneity on forgetting during personalized preference optimization, assuming that each user has a specific, definitive type of preference.

\section{Conclusion}
\vspace{-2pt}
This study explores the degree and nature of forgetting caused by personalized preference optimization in instruction-tuned LLMs. Our findings indicate a reduced likelihood of generating original responses, alongside a decrease in the generation of dispreferred responses. To address this, we introduce Base-Anchored Preference Optimization (BAPO), a method that anchors the likelihood of base responses during the preference optimization process. This approach effectively preserves global knowledge and general alignment while successfully accommodating personalized preferences. We have conducted extensive experiments to validate the efficacy of BAPO and its benefits.

\paragraph{Acknowledgements}
This work was supported by Institute of Information \& communications Technology Planning \& Evaluation (IITP) grant funded by the Korea government(MSIT) [No. RS-2024-00457882, AI Research Hub Project, 90\%]  and [No. 2019-0-00075, Artificial Intelligence Graduate School Program (KAIST), 10\%].

\paragraph{Limitations}
While BAPO effectively preserves existing knowledge by leveraging the base response, it is important to note that if the base model is biased, the fine-tuned personalized model may also exhibit a similar bias. This consideration is crucial for machine learning practitioners. In our experiments, we utilized Q-LoRA for fine-tuning. Although we conducted an ablation study varying model capacity by adjusting the LoRA rank, a full-finetune might show different tendencies. Nonetheless, using LoRA fine-tuning for personalization is a common approach in LLM context. Regarding computational costs, although BAPO requires the use of a base response, potentially increasing memory and computational demands during training, most of these costs can be mitigated by pre-generating the base responses and caching them as offline datasets. Concerning the anchoring strength hyperparameter, $\lambda$, the extent and scope of forgetting may vary based on the type of knowledge and the personalized preference types. Ideally, the $\lambda$ value should be adaptively assigned, but this aspect is left for future research.

\paragraph{Ethical Considerations} 
In developing and implementing our approach to align LLMs with personalized preferences, we must consider the potential implications. While BAPO aims to accommodate diverse user preferences, it is crucial to ensure this customization does not unintentionally reinforce harmful biases or perpetuate discrimination. Additionally, we must protect user privacy and data security, ensuring that personalization does not expose sensitive information or compromise user anonymity. Finally, maintaining a balance between personalized alignment and the integrity of general knowledge is essential to avoid scenarios where excessive personalization might result in misinformation or a loss of objective truth.

\paragraph{Use of AI Assistants} 
We utilize Copilot\footnote{\href{https://github.com/features/copilot}{https://github.com/features/copilot}} for the development of our code pipeline, primarily for its auto-completion capabilities. For drafting the paper, we employ ChatGPT\footnote{\href{https://chatgpt.com/}{https://chatgpt.com/}} to review the content, focusing on identifying grammatical errors and awkward expressions. However, the core content of the paper is original. We do not rely on AI assistants to generate any specific content that is closely related to the main claims of our research.

\bibliography{main}

\begin{thebibliography}{66}
\providecommand{\natexlab}[1]{#1}

\bibitem[{Abdin et~al.(2024)Abdin, Jacobs, Awan, Aneja, Awadallah, Awadalla, Bach, Bahree, Bakhtiari, Behl et~al.}]{phi3}
Marah Abdin, Sam~Ade Jacobs, Ammar~Ahmad Awan, Jyoti Aneja, Ahmed Awadallah, Hany Awadalla, Nguyen Bach, Amit Bahree, Arash Bakhtiari, Harkirat Behl, et~al. 2024.
\newblock Phi-3 technical report: A highly capable language model locally on your phone.
\newblock \emph{arXiv preprint arXiv:2404.14219}.

\bibitem[{Achiam et~al.(2023)Achiam, Adler, Agarwal, Ahmad, Akkaya, Aleman, Almeida, Altenschmidt, Altman, Anadkat et~al.}]{gpt4}
Josh Achiam, Steven Adler, Sandhini Agarwal, Lama Ahmad, Ilge Akkaya, Florencia~Leoni Aleman, Diogo Almeida, Janko Altenschmidt, Sam Altman, Shyamal Anadkat, et~al. 2023.
\newblock Gpt-4 technical report.
\newblock \emph{arXiv preprint arXiv:2303.08774}.

\bibitem[{Askell et~al.(2021)Askell, Bai, Chen, Drain, Ganguli, Henighan, Jones, Joseph, Mann, DasSarma et~al.}]{hhh_alilgnment}
Amanda Askell, Yuntao Bai, Anna Chen, Dawn Drain, Deep Ganguli, Tom Henighan, Andy Jones, Nicholas Joseph, Ben Mann, Nova DasSarma, et~al. 2021.
\newblock A general language assistant as a laboratory for alignment.
\newblock \emph{arXiv preprint arXiv:2112.00861}.

\bibitem[{Azar et~al.(2024)Azar, Guo, Piot, Munos, Rowland, Valko, and Calandriello}]{ipo}
Mohammad~Gheshlaghi Azar, Zhaohan~Daniel Guo, Bilal Piot, Remi Munos, Mark Rowland, Michal Valko, and Daniele Calandriello. 2024.
\newblock A general theoretical paradigm to understand learning from human preferences.
\newblock In \emph{International Conference on Artificial Intelligence and Statistics}, pages 4447--4455. PMLR.

\bibitem[{Bai et~al.(2022)Bai, Jones, Ndousse, Askell, Chen, DasSarma, Drain, Fort, Ganguli, Henighan et~al.}]{hh_rlhf}
Yuntao Bai, Andy Jones, Kamal Ndousse, Amanda Askell, Anna Chen, Nova DasSarma, Dawn Drain, Stanislav Fort, Deep Ganguli, Tom Henighan, et~al. 2022.
\newblock Training a helpful and harmless assistant with reinforcement learning from human feedback.
\newblock \emph{arXiv preprint arXiv:2204.05862}.

\bibitem[{Biderman et~al.(2024)Biderman, Ortiz, Portes, Paul, Greengard, Jennings, King, Havens, Chiley, Frankle et~al.}]{lora_forgets_less}
Dan Biderman, Jose~Gonzalez Ortiz, Jacob Portes, Mansheej Paul, Philip Greengard, Connor Jennings, Daniel King, Sam Havens, Vitaliy Chiley, Jonathan Frankle, et~al. 2024.
\newblock Lora learns less and forgets less.
\newblock \emph{arXiv preprint arXiv:2405.09673}.

\bibitem[{Bisk et~al.(2020)Bisk, Zellers, Gao, Choi et~al.}]{piqa}
Yonatan Bisk, Rowan Zellers, Jianfeng Gao, Yejin Choi, et~al. 2020.
\newblock Piqa: Reasoning about physical commonsense in natural language.
\newblock In \emph{Proceedings of the AAAI conference on artificial intelligence}, volume~34, pages 7432--7439.

\bibitem[{Bradley and Terry(1952)}]{bradley_terry}
Ralph~Allan Bradley and Milton~E Terry. 1952.
\newblock Rank analysis of incomplete block designs: I. the method of paired comparisons.
\newblock \emph{Biometrika}, 39(3/4):324--345.

\bibitem[{Chakraborty et~al.(2024)Chakraborty, Qiu, Yuan, Koppel, Huang, Manocha, Bedi, and Wang}]{maxmin_rlhf}
Souradip Chakraborty, Jiahao Qiu, Hui Yuan, Alec Koppel, Furong Huang, Dinesh Manocha, Amrit~Singh Bedi, and Mengdi Wang. 2024.
\newblock Maxmin-rlhf: Towards equitable alignment of large language models with diverse human preferences.
\newblock \emph{arXiv preprint arXiv:2402.08925}.

\bibitem[{Cheng et~al.(2023)Cheng, Xie, Bai, Dai, and Du}]{domain_specific_preference}
Pengyu Cheng, Jiawen Xie, Ke~Bai, Yong Dai, and Nan Du. 2023.
\newblock Everyone deserves a reward: Learning customized human preferences.
\newblock \emph{arXiv preprint arXiv:2309.03126}.

\bibitem[{Clark et~al.(2018)Clark, Cowhey, Etzioni, Khot, Sabharwal, Schoenick, and Tafjord}]{arc}
Peter Clark, Isaac Cowhey, Oren Etzioni, Tushar Khot, Ashish Sabharwal, Carissa Schoenick, and Oyvind Tafjord. 2018.
\newblock Think you have solved question answering? try arc, the ai2 reasoning challenge.
\newblock \emph{arXiv preprint arXiv:1803.05457}.

\bibitem[{Cui et~al.(2023)Cui, Yuan, Ding, Yao, Zhu, Ni, Xie, Liu, and Sun}]{ultrafeedback}
Ganqu Cui, Lifan Yuan, Ning Ding, Guanming Yao, Wei Zhu, Yuan Ni, Guotong Xie, Zhiyuan Liu, and Maosong Sun. 2023.
\newblock Ultrafeedback: Boosting language models with high-quality feedback.
\newblock \emph{arXiv preprint arXiv:2310.01377}.

\bibitem[{Dettmers et~al.(2024)Dettmers, Pagnoni, Holtzman, and Zettlemoyer}]{qlora}
Tim Dettmers, Artidoro Pagnoni, Ari Holtzman, and Luke Zettlemoyer. 2024.
\newblock Qlora: Efficient finetuning of quantized llms.
\newblock \emph{Advances in Neural Information Processing Systems}, 36.

\bibitem[{Dong et~al.(2023)Dong, Yuan, Lu, Li, Xue, Liu, Wang, Yuan, Zhou, and Zhou}]{llm_sft_data_composition}
Guanting Dong, Hongyi Yuan, Keming Lu, Chengpeng Li, Mingfeng Xue, Dayiheng Liu, Wei Wang, Zheng Yuan, Chang Zhou, and Jingren Zhou. 2023.
\newblock How abilities in large language models are affected by supervised fine-tuning data composition.
\newblock \emph{arXiv preprint arXiv:2310.05492}.

\bibitem[{Dou et~al.(2023)Dou, Zhou, Liu, Gao, Zhao, Shen, Zhou, Xi, Wang, Fan et~al.}]{loramoe}
Shihan Dou, Enyu Zhou, Yan Liu, Songyang Gao, Jun Zhao, Wei Shen, Yuhao Zhou, Zhiheng Xi, Xiao Wang, Xiaoran Fan, et~al. 2023.
\newblock The art of balancing: Revolutionizing mixture of experts for maintaining world knowledge in language model alignment.
\newblock \emph{arXiv preprint arXiv:2312.09979}.

\bibitem[{Gehrlein(1983)}]{condorcet_paradox1}
William~V Gehrlein. 1983.
\newblock Condorcet's paradox.
\newblock \emph{Theory and decision}, 15(2):161--197.

\bibitem[{Gehrlein(2002)}]{condorcet_paradox2}
William~V Gehrlein. 2002.
\newblock Condorcet's paradox and the likelihood of its occurrence: different perspectives on balanced preferences.
\newblock \emph{Theory and decision}, 52(2):171--199.

\bibitem[{Guo et~al.(2024)Guo, Cui, Yuan, Ding, Wang, Chen, Sun, Xie, Zhou, Lin et~al.}]{cpo_controllable}
Yiju Guo, Ganqu Cui, Lifan Yuan, Ning Ding, Jiexin Wang, Huimin Chen, Bowen Sun, Ruobing Xie, Jie Zhou, Yankai Lin, et~al. 2024.
\newblock Controllable preference optimization: Toward controllable multi-objective alignment.
\newblock \emph{arXiv preprint arXiv:2402.19085}.

\bibitem[{Hendrycks et~al.(2020)Hendrycks, Burns, Basart, Zou, Mazeika, Song, and Steinhardt}]{mmlu}
Dan Hendrycks, Collin Burns, Steven Basart, Andy Zou, Mantas Mazeika, Dawn Song, and Jacob Steinhardt. 2020.
\newblock Measuring massive multitask language understanding.
\newblock \emph{arXiv preprint arXiv:2009.03300}.

\bibitem[{Hong et~al.(2024)Hong, Lee, and Thorne}]{orpo}
Jiwoo Hong, Noah Lee, and James Thorne. 2024.
\newblock Reference-free monolithic preference optimization with odds ratio.
\newblock \emph{arXiv preprint arXiv:2403.07691}.

\bibitem[{Hu et~al.(2021)Hu, Shen, Wallis, Allen-Zhu, Li, Wang, Wang, and Chen}]{lora}
Edward~J Hu, Yelong Shen, Phillip Wallis, Zeyuan Allen-Zhu, Yuanzhi Li, Shean Wang, Lu~Wang, and Weizhu Chen. 2021.
\newblock Lora: Low-rank adaptation of large language models.
\newblock \emph{arXiv preprint arXiv:2106.09685}.

\bibitem[{Jafari et~al.(2024)Jafari, Mekala, Yu, and Berg-Kirkpatrick}]{morl_prompt}
Yasaman Jafari, Dheeraj Mekala, Rose Yu, and Taylor Berg-Kirkpatrick. 2024.
\newblock Morl-prompt: An empirical analysis of multi-objective reinforcement learning for discrete prompt optimization.
\newblock \emph{arXiv preprint arXiv:2402.11711}.

\bibitem[{Jang et~al.(2023)Jang, Kim, Lin, Wang, Hessel, Zettlemoyer, Hajishirzi, Choi, and Ammanabrolu}]{psoups}
Joel Jang, Seungone Kim, Bill~Yuchen Lin, Yizhong Wang, Jack Hessel, Luke Zettlemoyer, Hannaneh Hajishirzi, Yejin Choi, and Prithviraj Ammanabrolu. 2023.
\newblock Personalized soups: Personalized large language model alignment via post-hoc parameter merging.
\newblock \emph{arXiv preprint arXiv:2310.11564}.

\bibitem[{Ji et~al.(2023)Ji, Qiu, Chen, Zhang, Lou, Wang, Duan, He, Zhou, Zhang et~al.}]{alignment_survey}
Jiaming Ji, Tianyi Qiu, Boyuan Chen, Borong Zhang, Hantao Lou, Kaile Wang, Yawen Duan, Zhonghao He, Jiayi Zhou, Zhaowei Zhang, et~al. 2023.
\newblock Ai alignment: A comprehensive survey.
\newblock \emph{arXiv preprint arXiv:2310.19852}.

\bibitem[{Jiang et~al.(2024)Jiang, Sablayrolles, Roux, Mensch, Savary, Bamford, Chaplot, Casas, Hanna, Bressand et~al.}]{mixtral}
Albert~Q Jiang, Alexandre Sablayrolles, Antoine Roux, Arthur Mensch, Blanche Savary, Chris Bamford, Devendra~Singh Chaplot, Diego de~las Casas, Emma~Bou Hanna, Florian Bressand, et~al. 2024.
\newblock Mixtral of experts.
\newblock \emph{arXiv preprint arXiv:2401.04088}.

\bibitem[{Jin and Ren(2024)}]{forecasting_forgetting_llms}
Xisen Jin and Xiang Ren. 2024.
\newblock What will my model forget? forecasting forgotten examples in language model refinement.
\newblock \emph{arXiv preprint arXiv:2402.01865}.

\bibitem[{Lee et~al.(2023)Lee, Phatale, Mansoor, Lu, Mesnard, Bishop, Carbune, and Rastogi}]{rlaif}
Harrison Lee, Samrat Phatale, Hassan Mansoor, Kellie Lu, Thomas Mesnard, Colton Bishop, Victor Carbune, and Abhinav Rastogi. 2023.
\newblock Rlaif: Scaling reinforcement learning from human feedback with ai feedback.
\newblock \emph{arXiv preprint arXiv:2309.00267}.

\bibitem[{Li et~al.(2024)Li, Lipton, and Leqi}]{p_rlhf}
Xinyu Li, Zachary~C Lipton, and Liu Leqi. 2024.
\newblock Personalized language modeling from personalized human feedback.
\newblock \emph{arXiv preprint arXiv:2402.05133}.

\bibitem[{Lin et~al.(2023)Lin, Tan, Lin, Zheng, Pi, Zhang, Diao, Wang, Zhao, Yao et~al.}]{mitigating_alignment_tax}
Yong Lin, Lu~Tan, Hangyu Lin, Zeming Zheng, Renjie Pi, Jipeng Zhang, Shizhe Diao, Haoxiang Wang, Han Zhao, Yuan Yao, et~al. 2023.
\newblock Speciality vs generality: An empirical study on catastrophic forgetting in fine-tuning foundation models.
\newblock \emph{arXiv preprint arXiv:2309.06256}.

\bibitem[{Liu et~al.(2023)Liu, Zhao, Joshi, Khalman, Saleh, Liu, and Liu}]{rso}
Tianqi Liu, Yao Zhao, Rishabh Joshi, Misha Khalman, Mohammad Saleh, Peter~J Liu, and Jialu Liu. 2023.
\newblock Statistical rejection sampling improves preference optimization.
\newblock \emph{arXiv preprint arXiv:2309.06657}.

\bibitem[{Loshchilov and Hutter(2016)}]{cosine}
Ilya Loshchilov and Frank Hutter. 2016.
\newblock Sgdr: Stochastic gradient descent with warm restarts.
\newblock \emph{arXiv preprint arXiv:1608.03983}.

\bibitem[{Lu et~al.(2024)Lu, Yu, Huang, Fan, Lin, and Zhou}]{online_merging_rlhf}
Keming Lu, Bowen Yu, Fei Huang, Yang Fan, Runji Lin, and Chang Zhou. 2024.
\newblock Online merging optimizers for boosting rewards and mitigating tax in alignment.
\newblock \emph{arXiv preprint arXiv:2405.17931}.

\bibitem[{Lu et~al.(2022)Lu, Mishra, Xia, Qiu, Chang, Zhu, Tafjord, Clark, and Kalyan}]{science_qa}
Pan Lu, Swaroop Mishra, Tanglin Xia, Liang Qiu, Kai-Wei Chang, Song-Chun Zhu, Oyvind Tafjord, Peter Clark, and Ashwin Kalyan. 2022.
\newblock Learn to explain: Multimodal reasoning via thought chains for science question answering.
\newblock \emph{Advances in Neural Information Processing Systems}, 35:2507--2521.

\bibitem[{Luo et~al.(2023)Luo, Yang, Meng, Li, Zhou, and Zhang}]{empirical_continual_finetuning}
Yun Luo, Zhen Yang, Fandong Meng, Yafu Li, Jie Zhou, and Yue Zhang. 2023.
\newblock An empirical study of catastrophic forgetting in large language models during continual fine-tuning.
\newblock \emph{arXiv preprint arXiv:2308.08747}.

\bibitem[{McCloskey and Cohen(1989)}]{cl2}
Michael McCloskey and Neal~J Cohen. 1989.
\newblock Catastrophic interference in connectionist networks: The sequential learning problem.
\newblock In \emph{Psychology of learning and motivation}, volume~24, pages 109--165. Elsevier.

\bibitem[{Mermillod et~al.(2013)Mermillod, Bugaiska, and Bonin}]{stability1}
Martial Mermillod, Aur{\'e}lia Bugaiska, and Patrick Bonin. 2013.
\newblock The stability-plasticity dilemma: Investigating the continuum from catastrophic forgetting to age-limited learning effects.

\bibitem[{Meta(2024)}]{llama3}
Meta. 2024.
\newblock \href {https://about.fb.com/news/2024/04/meta-ai-assistant-built-with-llama-3/} {Introducing meta llama 3: The most capable openly available llm to date}.

\bibitem[{Munos et~al.(2023)Munos, Valko, Calandriello, Azar, Rowland, Guo, Tang, Geist, Mesnard, Michi et~al.}]{nash_human_feedback}
R{\'e}mi Munos, Michal Valko, Daniele Calandriello, Mohammad~Gheshlaghi Azar, Mark Rowland, Zhaohan~Daniel Guo, Yunhao Tang, Matthieu Geist, Thomas Mesnard, Andrea Michi, et~al. 2023.
\newblock Nash learning from human feedback.
\newblock \emph{arXiv preprint arXiv:2312.00886}.

\bibitem[{Ouyang et~al.(2022)Ouyang, Wu, Jiang, Almeida, Wainwright, Mishkin, Zhang, Agarwal, Slama, Ray et~al.}]{rlhf}
Long Ouyang, Jeffrey Wu, Xu~Jiang, Diogo Almeida, Carroll Wainwright, Pamela Mishkin, Chong Zhang, Sandhini Agarwal, Katarina Slama, Alex Ray, et~al. 2022.
\newblock Training language models to follow instructions with human feedback.
\newblock \emph{Advances in neural information processing systems}, 35:27730--27744.

\bibitem[{Pal et~al.(2024)Pal, Karkhanis, Dooley, Roberts, Naidu, and White}]{smaug_dpop}
Arka Pal, Deep Karkhanis, Samuel Dooley, Manley Roberts, Siddartha Naidu, and Colin White. 2024.
\newblock Smaug: Fixing failure modes of preference optimisation with dpo-positive.
\newblock \emph{arXiv preprint arXiv:2402.13228}.

\bibitem[{Parisi et~al.(2019)Parisi, Kemker, Part, Kanan, and Wermter}]{cl1}
German~I Parisi, Ronald Kemker, Jose~L Part, Christopher Kanan, and Stefan Wermter. 2019.
\newblock Continual lifelong learning with neural networks: A review.
\newblock \emph{Neural networks}, 113:54--71.

\bibitem[{Park et~al.(2024)Park, Liu, Zhang, and Ozdaglar}]{personalization_preference_aggregation}
Chanwoo Park, Mingyang Liu, Kaiqing Zhang, and Asuman Ozdaglar. 2024.
\newblock Principled rlhf from heterogeneous feedback via personalization and preference aggregation.
\newblock \emph{arXiv preprint arXiv:2405.00254}.

\bibitem[{Peng et~al.(2023)Peng, Li, He, Galley, and Gao}]{alpaca_gpt4}
Baolin Peng, Chunyuan Li, Pengcheng He, Michel Galley, and Jianfeng Gao. 2023.
\newblock Instruction tuning with gpt-4.
\newblock \emph{arXiv preprint arXiv:2304.03277}.

\bibitem[{Rafailov et~al.(2024)Rafailov, Sharma, Mitchell, Manning, Ermon, and Finn}]{dpo}
Rafael Rafailov, Archit Sharma, Eric Mitchell, Christopher~D Manning, Stefano Ermon, and Chelsea Finn. 2024.
\newblock Direct preference optimization: Your language model is secretly a reward model.
\newblock \emph{Advances in Neural Information Processing Systems}, 36.

\bibitem[{Rame et~al.(2024)Rame, Couairon, Dancette, Gaya, Shukor, Soulier, and Cord}]{reward_soups}
Alexandre Rame, Guillaume Couairon, Corentin Dancette, Jean-Baptiste Gaya, Mustafa Shukor, Laure Soulier, and Matthieu Cord. 2024.
\newblock Rewarded soups: towards pareto-optimal alignment by interpolating weights fine-tuned on diverse rewards.
\newblock \emph{Advances in Neural Information Processing Systems}, 36.

\bibitem[{Sap et~al.(2019)Sap, Rashkin, Chen, LeBras, and Choi}]{social_i_qa}
Maarten Sap, Hannah Rashkin, Derek Chen, Ronan LeBras, and Yejin Choi. 2019.
\newblock Socialiqa: Commonsense reasoning about social interactions.
\newblock \emph{arXiv preprint arXiv:1904.09728}.

\bibitem[{Schulman et~al.(2017)Schulman, Wolski, Dhariwal, Radford, and Klimov}]{ppo}
John Schulman, Filip Wolski, Prafulla Dhariwal, Alec Radford, and Oleg Klimov. 2017.
\newblock Proximal policy optimization algorithms.
\newblock \emph{arXiv preprint arXiv:1707.06347}.

\bibitem[{Stiennon et~al.(2020)Stiennon, Ouyang, Wu, Ziegler, Lowe, Voss, Radford, Amodei, and Christiano}]{learning_to_summarize_rlhf}
Nisan Stiennon, Long Ouyang, Jeffrey Wu, Daniel Ziegler, Ryan Lowe, Chelsea Voss, Alec Radford, Dario Amodei, and Paul~F Christiano. 2020.
\newblock Learning to summarize with human feedback.
\newblock \emph{Advances in Neural Information Processing Systems}, 33:3008--3021.

\bibitem[{Talmor et~al.(2018)Talmor, Herzig, Lourie, and Berant}]{commonsense_qa}
Alon Talmor, Jonathan Herzig, Nicholas Lourie, and Jonathan Berant. 2018.
\newblock Commonsenseqa: A question answering challenge targeting commonsense knowledge.
\newblock \emph{arXiv preprint arXiv:1811.00937}.

\bibitem[{Taori et~al.(2023)Taori, Gulrajani, Zhang, Dubois, Li, Guestrin, Liang, and Hashimoto}]{alpaca}
Rohan Taori, Ishaan Gulrajani, Tianyi Zhang, Yann Dubois, Xuechen Li, Carlos Guestrin, Percy Liang, and Tatsunori~B. Hashimoto. 2023.
\newblock Stanford alpaca: An instruction-following llama model.
\newblock \url{https://github.com/tatsu-lab/stanford_alpaca}.

\bibitem[{Team et~al.(2024)Team, Mesnard, Hardin, Dadashi, Bhupatiraju, Pathak, Sifre, Rivi{\`e}re, Kale, Love et~al.}]{gemma}
Gemma Team, Thomas Mesnard, Cassidy Hardin, Robert Dadashi, Surya Bhupatiraju, Shreya Pathak, Laurent Sifre, Morgane Rivi{\`e}re, Mihir~Sanjay Kale, Juliette Love, et~al. 2024.
\newblock Gemma: Open models based on gemini research and technology.
\newblock \emph{arXiv preprint arXiv:2403.08295}.

\bibitem[{Touvron et~al.(2023)Touvron, Martin, Stone, Albert, Almahairi, Babaei, Bashlykov, Batra, Bhargava, Bhosale et~al.}]{llama2}
Hugo Touvron, Louis Martin, Kevin Stone, Peter Albert, Amjad Almahairi, Yasmine Babaei, Nikolay Bashlykov, Soumya Batra, Prajjwal Bhargava, Shruti Bhosale, et~al. 2023.
\newblock Llama 2: Open foundation and fine-tuned chat models.
\newblock \emph{arXiv preprint arXiv:2307.09288}.

\bibitem[{Tunstall et~al.(2023{\natexlab{a}})Tunstall, Beeching, Lambert, Rajani, Huang, Rasul, Rush, and Wolf}]{alignment_handbook}
Lewis Tunstall, Edward Beeching, Nathan Lambert, Nazneen Rajani, Shengyi Huang, Kashif Rasul, Alexander~M. Rush, and Thomas Wolf. 2023{\natexlab{a}}.
\newblock The alignment handbook.
\newblock \url{https://github.com/huggingface/alignment-handbook}.

\bibitem[{Tunstall et~al.(2023{\natexlab{b}})Tunstall, Beeching, Lambert, Rajani, Rasul, Belkada, Huang, von Werra, Fourrier, Habib et~al.}]{zephyr}
Lewis Tunstall, Edward Beeching, Nathan Lambert, Nazneen Rajani, Kashif Rasul, Younes Belkada, Shengyi Huang, Leandro von Werra, Cl{\'e}mentine Fourrier, Nathan Habib, et~al. 2023{\natexlab{b}}.
\newblock Zephyr: Direct distillation of lm alignment.
\newblock \emph{arXiv preprint arXiv:2310.16944}.

\bibitem[{Wang et~al.(2024{\natexlab{a}})Wang, Lin, Xiong, Yang, Diao, Qiu, Zhao, and Zhang}]{dpa_morl}
Haoxiang Wang, Yong Lin, Wei Xiong, Rui Yang, Shizhe Diao, Shuang Qiu, Han Zhao, and Tong Zhang. 2024{\natexlab{a}}.
\newblock Arithmetic control of llms for diverse user preferences: Directional preference alignment with multi-objective rewards.
\newblock \emph{arXiv preprint arXiv:2402.18571}.

\bibitem[{Wang et~al.(2024{\natexlab{b}})Wang, Zhang, Su, and Zhu}]{cl_survey}
Liyuan Wang, Xingxing Zhang, Hang Su, and Jun Zhu. 2024{\natexlab{b}}.
\newblock A comprehensive survey of continual learning: Theory, method and application.
\newblock \emph{IEEE Transactions on Pattern Analysis and Machine Intelligence}.

\bibitem[{Wang et~al.(2023{\natexlab{a}})Wang, Li, Chen, Song, Lin, Cao, Liu, and Sui}]{better_reasoners_alignment}
Peiyi Wang, Lei Li, Liang Chen, Feifan Song, Binghuai Lin, Yunbo Cao, Tianyu Liu, and Zhifang Sui. 2023{\natexlab{a}}.
\newblock Making large language models better reasoners with alignment.
\newblock \emph{arXiv preprint arXiv:2309.02144}.

\bibitem[{Wang et~al.(2023{\natexlab{b}})Wang, Yang, Shen, and Huang}]{stability2}
Zhenyi Wang, Enneng Yang, Li~Shen, and Heng Huang. 2023{\natexlab{b}}.
\newblock A comprehensive survey of forgetting in deep learning beyond continual learning.
\newblock \emph{arXiv preprint arXiv:2307.09218}.

\bibitem[{Wang et~al.(2023{\natexlab{c}})Wang, Yang, Shen, and Huang}]{forgetting_survey}
Zhenyi Wang, Enneng Yang, Li~Shen, and Heng Huang. 2023{\natexlab{c}}.
\newblock A comprehensive survey of forgetting in deep learning beyond continual learning.
\newblock \emph{arXiv preprint arXiv:2307.09218}.

\bibitem[{Wolf et~al.(2019)Wolf, Debut, Sanh, Chaumond, Delangue, Moi, Cistac, Rault, Louf, Funtowicz et~al.}]{hf-transformers}
Thomas Wolf, Lysandre Debut, Victor Sanh, Julien Chaumond, Clement Delangue, Anthony Moi, Pierric Cistac, Tim Rault, R{\'e}mi Louf, Morgan Funtowicz, et~al. 2019.
\newblock Huggingface's transformers: State-of-the-art natural language processing.
\newblock \emph{arXiv preprint arXiv:1910.03771}.

\bibitem[{Xu et~al.(2024)Xu, Sharaf, Chen, Tan, Shen, Van~Durme, Murray, and Kim}]{cpo_contrastive}
Haoran Xu, Amr Sharaf, Yunmo Chen, Weiting Tan, Lingfeng Shen, Benjamin Van~Durme, Kenton Murray, and Young~Jin Kim. 2024.
\newblock Contrastive preference optimization: Pushing the boundaries of llm performance in machine translation.
\newblock \emph{arXiv preprint arXiv:2401.08417}.

\bibitem[{Yang et~al.(2024)Yang, Pan, Luo, Qiu, Zhong, Yu, and Chen}]{rewards_in_context}
Rui Yang, Xiaoman Pan, Feng Luo, Shuang Qiu, Han Zhong, Dong Yu, and Jianshu Chen. 2024.
\newblock Rewards-in-context: Multi-objective alignment of foundation models with dynamic preference adjustment.
\newblock \emph{arXiv preprint arXiv:2402.10207}.

\bibitem[{Zeng et~al.(2023)Zeng, Dai, Cheng, Hu, Chen, Du, and Xu}]{more}
Dun Zeng, Yong Dai, Pengyu Cheng, Tianhao Hu, Wanshun Chen, Nan Du, and Zenglin Xu. 2023.
\newblock On diversified preferences of large language model alignment.
\newblock \emph{arXiv preprint arXiv:2312.07401}.

\bibitem[{Zheng et~al.(2023)Zheng, Dou, Gao, Hua, Shen, Wang, Liu, Jin, Liu, Zhou et~al.}]{scerets_rlhf_ppo}
Rui Zheng, Shihan Dou, Songyang Gao, Yuan Hua, Wei Shen, Binghai Wang, Yan Liu, Senjie Jin, Qin Liu, Yuhao Zhou, et~al. 2023.
\newblock Secrets of rlhf in large language models part i: Ppo.
\newblock \emph{arXiv preprint arXiv:2307.04964}.

\bibitem[{Zhong et~al.(2024)Zhong, Ma, Zhang, Yang, Zhang, Qi, and Yang}]{panacea}
Yifan Zhong, Chengdong Ma, Xiaoyuan Zhang, Ziran Yang, Qingfu Zhang, Siyuan Qi, and Yaodong Yang. 2024.
\newblock Panacea: Pareto alignment via preference adaptation for llms.
\newblock \emph{arXiv preprint arXiv:2402.02030}.

\bibitem[{Zhu et~al.(2023)Zhu, Jordan, and Jiao}]{zhu2023principled}
Banghua Zhu, Michael Jordan, and Jiantao Jiao. 2023.
\newblock Principled reinforcement learning with human feedback from pairwise or k-wise comparisons.
\newblock In \emph{International Conference on Machine Learning}, pages 43037--43067. PMLR.

\end{thebibliography}

\clearpage
\appendix
\clearpage
\setcounter{page}{1}
\onecolumn

\appendix
\section{Dataset details}
\label{appendix:datasets}
In our experiments, we utilize two datasets written in English for personalized preference optimization: the DSP (Domain-Specific Preference) and P-Soups datasets. The DSP dataset, proposed by \citep{domain_specific_preference}, comprises 13,000 prompts selected from the 52,000 Alpaca datasets \citep{alpaca}. It features preferred responses tailored to specific queries across four practical domains: Academy, Business, Entertainment, and Literature \& Art. Each prompt includes five responses: one from each of the four domains and the original Alpaca response. In our pairwise preference setup, the response from the corresponding domain is selected as the preferred one, while the others are considered rejected. This results in 52,000 pairs per dataset, with 2,000 pairs designated for the validation split. The P-Soups dataset \citep{psoups}, simulated by GPT-4, consists of pairwise feedback data where the AI is instructed to choose the better of two candidate responses. This dataset builds on prompt instances from Alpaca-GPT4, with additional prompts provided to ensure consistency in preference criteria. To facilitate experiments that require the same prompt across different preference types, we exclude the additional prompts introduced by P-Soups, keeping only the original prompts from the Alpaca dataset. The P-Soups dataset categorizes six conflicting preferences into three dimensions: expertise, informativeness, and friendliness, resulting in six preference combinations. Out of the 10,000 Alpaca-GPT4 prompts \citep{alpaca_gpt4}, we derive between 47,000 to 49,000 pairwise feedback entries per preference type, allocating 45,000 samples for the training split and the remainder for the validation split. Note that while response lengths may vary among preferences in the P-Soups dataset, they remain fairly consistent across different domains in the DSP dataset.



\section{Resources \& Others} We use six RTX A6000 48GB GPU cards for our experiments, although we do not employ multi-GPU training. The GPU hours required for each run vary, but typically, running one epoch on the Phi-3-mini model with a LoRA rank of 32, including time for evaluation, takes about 10 hours. To reproduce all the results in this paper, approximately 1,600 GPU hours are required. We employ the Hugging Face Transformers library \citep{hf-transformers} for the overall code.

\section{Changes in Log Probabilities of Responses}
\label{appendix:log_probs_responses}

\begin{figure}[ht!]
    \centering
    \includegraphics[width=0.9\textwidth]{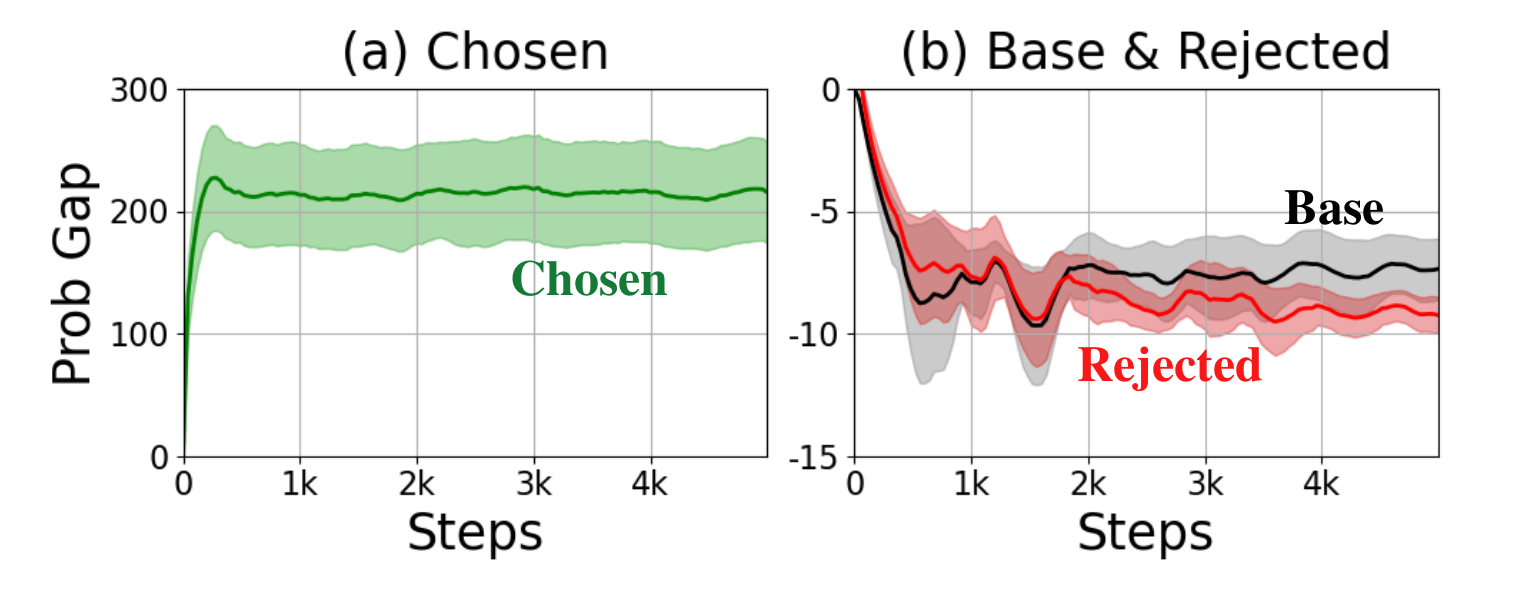}
    \caption{Log probabilities gap for responses on \textbf{P-Soups}\,\citep{psoups} datasets.}
    \label{fig:logps_gap_psoups}
\end{figure}

\begin{figure}[ht!]
    \centering
    \includegraphics[width=0.9\textwidth]{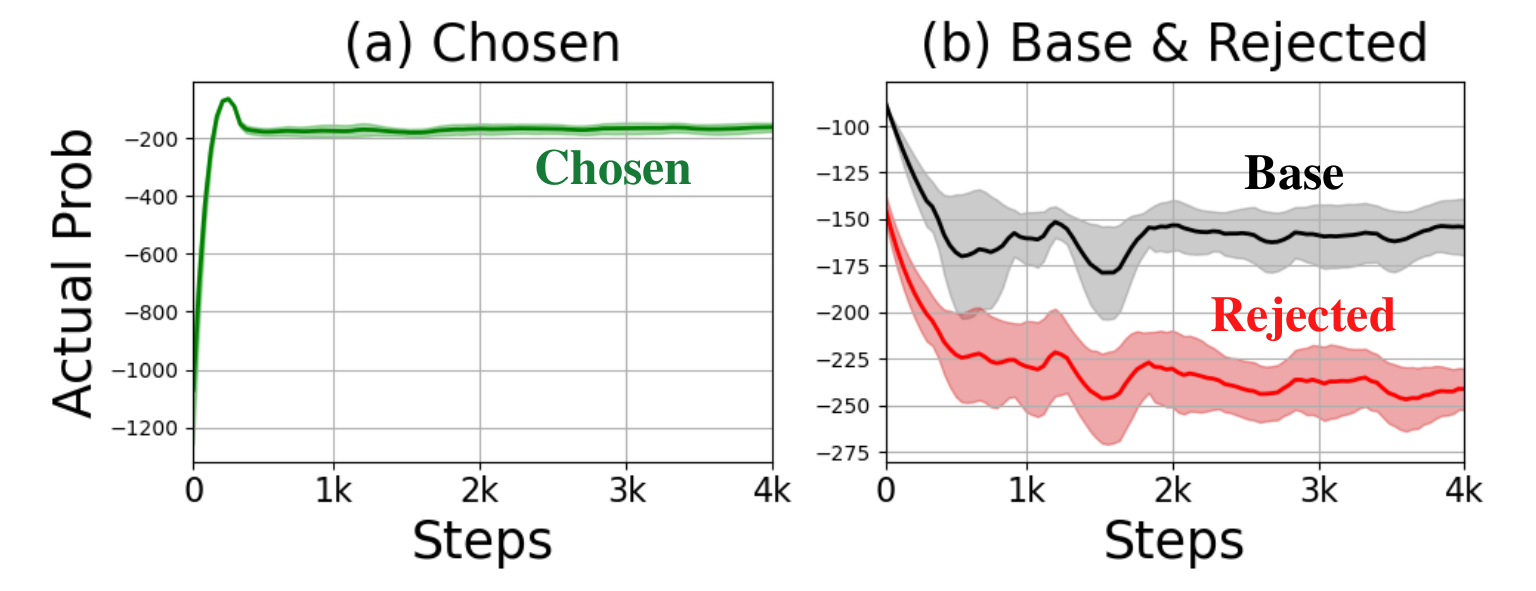}
    \caption{Actual log probabilities for responses on \textbf{P-Soups}\,\citep{psoups} datasets.}
    \label{fig:logps_actual_psoups}
\end{figure}

In \autoref{fig:logps_gap_psoups}, we show the change in the log probability gap during personalized preference optimization on the P-Soups datasets. The gap indicates how the log probabilities of each response differ compared to the initial frozen reference model. During the early stages of learning, the personalized preference optimization rapidly increases the gap for the chosen response, which then quickly saturates. Meanwhile, the log probabilities for the rejected and base responses decrease rapidly. \autoref{fig:logps_actual_psoups} displays the actual log probabilities for each response.


\begin{figure}[ht!]
    \centering
    \includegraphics[width=0.9\textwidth]{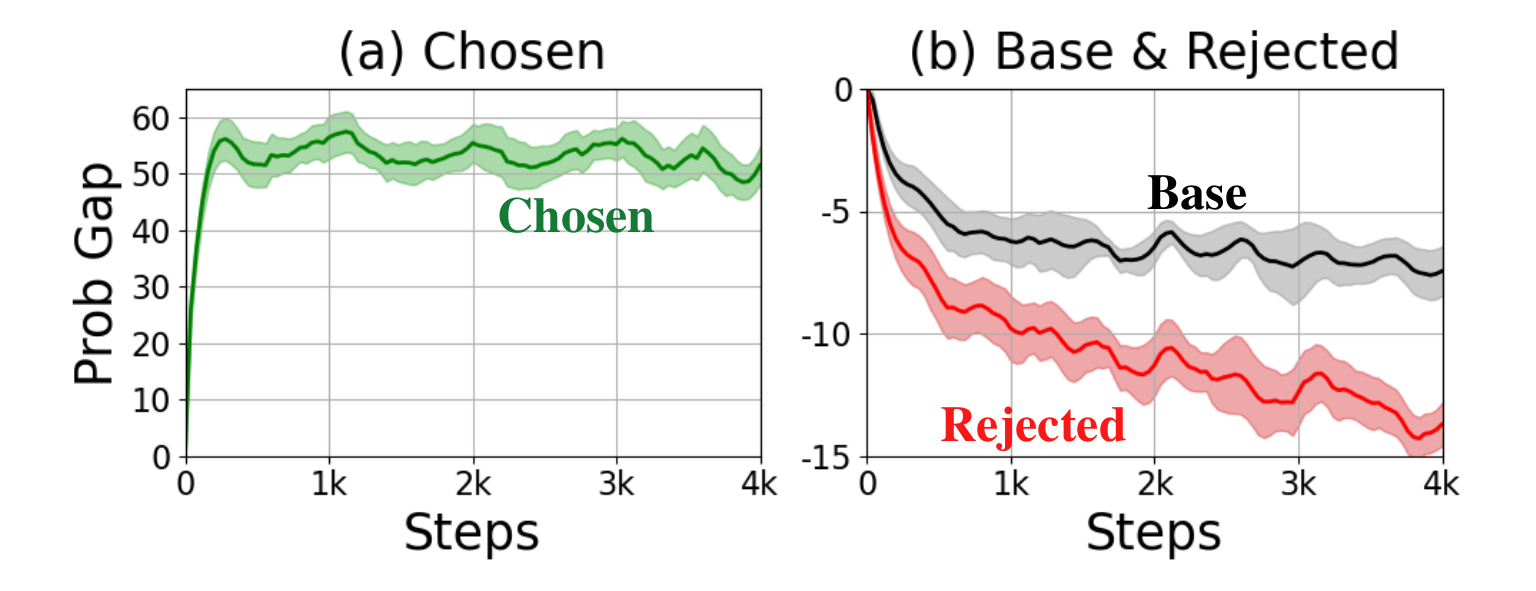}
    \caption{Log probabilities gap for responses on \textbf{DSP}\,\citep{domain_specific_preference} datasets.}
    \label{fig:logps_gap_dsp}
\end{figure}

\begin{figure}[ht!]
    \centering
    \includegraphics[width=0.9\textwidth]{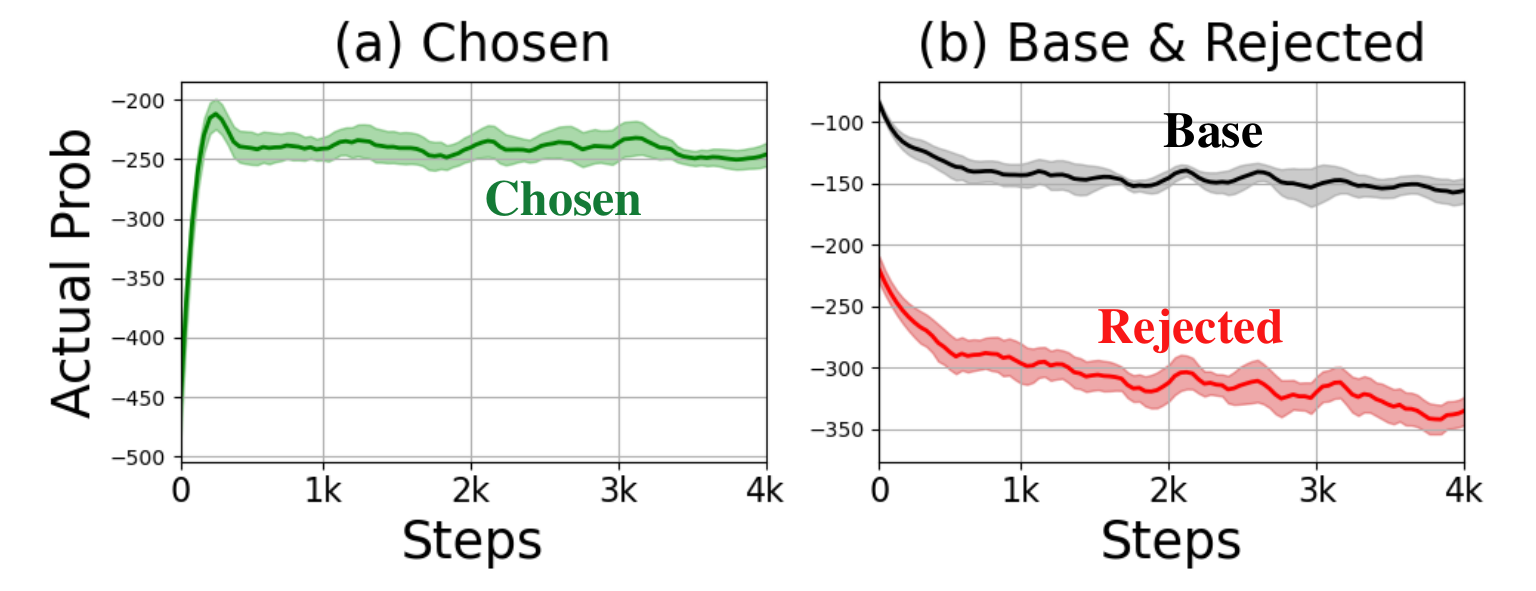}
    \caption{Actual log probabilities for responses on \textbf{P-Soups}\,\citep{psoups} datasets.}
    \label{fig:logps_actual_dsp}
\end{figure}

In \autoref{fig:logps_gap_dsp}, we present the same analysis on the DSP datasets, with the actual log probability plot in \autoref{fig:logps_actual_dsp}. While the scale differs, the overall trend is consistent with the P-Soups experiment. The main distinction is that P-Soups datasets\,(style preference) significantly impacts the log probabilities of the base response more than the DSP datasets\,(domain preference). We conjecture that the DSP datasets aim to differentiate responses across different domains, whereas P-Soups datasets share the same domain but differ only in superficial style, thus having a greater effect on the base response.

\clearpage
\section{Proof of Proposition 1}
\label{appendix:proof}
First, observe that under the linear reward model assumption, maximum likelihood estimation (MLE) of unknown parameter $\theta_{\star}$ is obtained as follows:
\begin{align}
\label{eq:MLE_estimation of BT model}
\hat{\theta}_{\text{MLE}}=\text{argmin}_{\theta\in\mathcal{\theta}_{B}}\mathcal{L}_{\text{BT}}(\theta)
= \text{argmin}_{\theta\in\mathcal{\theta}_{B}} \sum_{i=1}^{n}  -\log \left( \sigma \left( \langle \theta, \phi(x^i, y_w^i) - \phi(x^i, y_l^i) \rangle \right) \right),
\end{align}
where, $\sigma$ is a sigmoid function and $(x^i, y_{w}^{i},  y_{l}^{i})$ denotes $i$-th (out of n) preference sample with context $x^i$, winning and losing response $y_{w}^{i},  y_{l}^{i}$ respectively. Also, and $\mathcal{\theta}_{B} = \{\theta \in \mathbb{R}^d : \|\theta\| \leq B\}$.\\

In order to prove the \textbf{Proposition~\ref{prop:sample_complexity}}, we bring the latest result for the sample complexity bound for the linear preference model which is provided in the \textbf{Lemma 3.1} of~\citep{zhu2023principled}. We restate the lemma here for the completeness.

\begin{lemma}
\label{lem:sample_complexity_zhu}
Assume, $\phi(y,x)\leq L$ for all possible response, context pairs $(y,x)$ and $\theta_{\star}\leq B$ is unknown parameter for linear model in eq.~\ref{eq:MLE_estimation of BT model}. Then, for $\lambda>0$, constant $C'>0$ and estimator $\hat{\theta}_{MLE}$ for the Bradley-Terry model loss (eq.~\ref{eq:MLE_estimation of BT model}), the following confidence bound holds with probability $1-\delta$:
\begin{align}
\|\hat{\theta}_{MLE} - \theta_{\star}\|_{\Sigma_{D}+\lambda I} \leq C' \cdot \sqrt{\frac{d + \log(1/\delta)}{\gamma^2 n} + \lambda B^2}.
\end{align}
Here, $\Sigma_D = \frac{1}{n} \sum_{i=1}^n (\phi(x^i, y_{w}^{i}) - \phi(x^i, y_{l}^{i}))(\phi(x^i, y_{w}^{i})-\phi(x^i, y_{l}^{i}))^\top$, $\gamma = 1/(2 + \exp(-LB) + \exp(LB))$.\\
\end{lemma}

Now, suppose we have full information of $\theta_{\star}^{G}$ and without loss of generality, $\theta_{\star}^{G}$ has nonzero values on the first d-k dimensions. (Note that according to \textbf{Assumption.~\ref{assumption2:linear_model_independence}}, this automatically implies that $\theta_{\star}^{L}$ has nonzero components only on the last k dimensions.) Here, for the ease of analysis, we truncate only nonzero parts of $\theta_{\star}^{L}$ 
to make it a k-dimensional vector. Also, denote $\phi^{L}(y,x)=\phi(y,x)_{d-k+1:d}$ be the last k dimensional part of feature vector that governs personalization reward $L$. With this, we can calculate MLE of BT model only for the last k dimensional components in the following way:
\begin{align}
\label{eq:MLE_estimation of BT model_last_k_dim}
\hat{\theta}_{\text{MLE}}^{L}=\text{argmin}_{\theta^L\in\mathcal{\theta}_{B'}}\mathcal{L}_{\text{BT}}(\theta^L)
= \text{argmin}_{\theta^L\in\mathcal{\theta}_{B'}} \sum_{i=1}^{n}  -\log \left( \sigma \left( \langle \theta^{L}, \phi^{L}(x^i, y_w^i) - \phi^{L}(x^i, y_l^i) \rangle \right) \right),
\end{align}
where $\mathcal{\theta}_{B'} = \{\theta \in \mathbb{R}^k : \|\theta\| \leq B'\}$.\\

Now,
with \textbf{Lemma.~\ref{lem:sample_complexity_zhu}} and \textbf{Assumptions~\ref{assumption1:existence_of_utilities},~\ref{assumption2:linear_model_independence}}, it is easy to see the following confidence bound for the $\theta_{\star}^{L}$ holds by the following lemma.
\begin{lemma}
\label{lem:sample_complexity_personalize}
With $\phi(y,x)_{d-k+1:d}\leq L'$ for all possible response, context pairs $(y,x)$ and $\theta_{\star}^{L}\leq B'$. Then, for $\lambda'>0$, constant $C''>0$ and unknown parameter $\hat{\theta}^{L}_{MLE}$ from the modified Bradley-Terry model loss (eq.~\ref{eq:MLE_estimation of BT model_last_k_dim}), the following confidence bound holds with probability $1-\delta$:
\begin{align}
\|\hat{\theta}_{MLE}^{L} - \theta_{\star}^{L}\|_{\Sigma_{D}^{L}+\lambda' I} \leq C'' \cdot \sqrt{\frac{d + \log(1/\delta)}{{\gamma'}^2 n} + \lambda' {B'}^2}.
\end{align}
Here, $\Sigma_{D}^{L} = \frac{1}{n} \sum_{i=1}^n (\phi^{L}(x^i, y_{w}^{i}) - \phi^{L}(x^i, y_{l}^{i}))(\phi^{L}(x^i, y_{w}^{i})-\phi^{L}(x^i, y_{l}^{i}))^\top$, $\gamma' = 1/(2 + \exp(-L'B') + \exp(L'B')).$\\
\end{lemma}

Combining \textbf{Lemma 1} and \textbf{Lemma 2}, we see that \textbf{Proposition.~\ref{prop:sample_complexity}} holds. 

\clearpage
\section{Complete Example Responses of \autoref{fig:main_bapo}}
\label{appendix:main_full_example}

\begin{figure}[ht!]
    \centering
    \includegraphics[width=\textwidth]{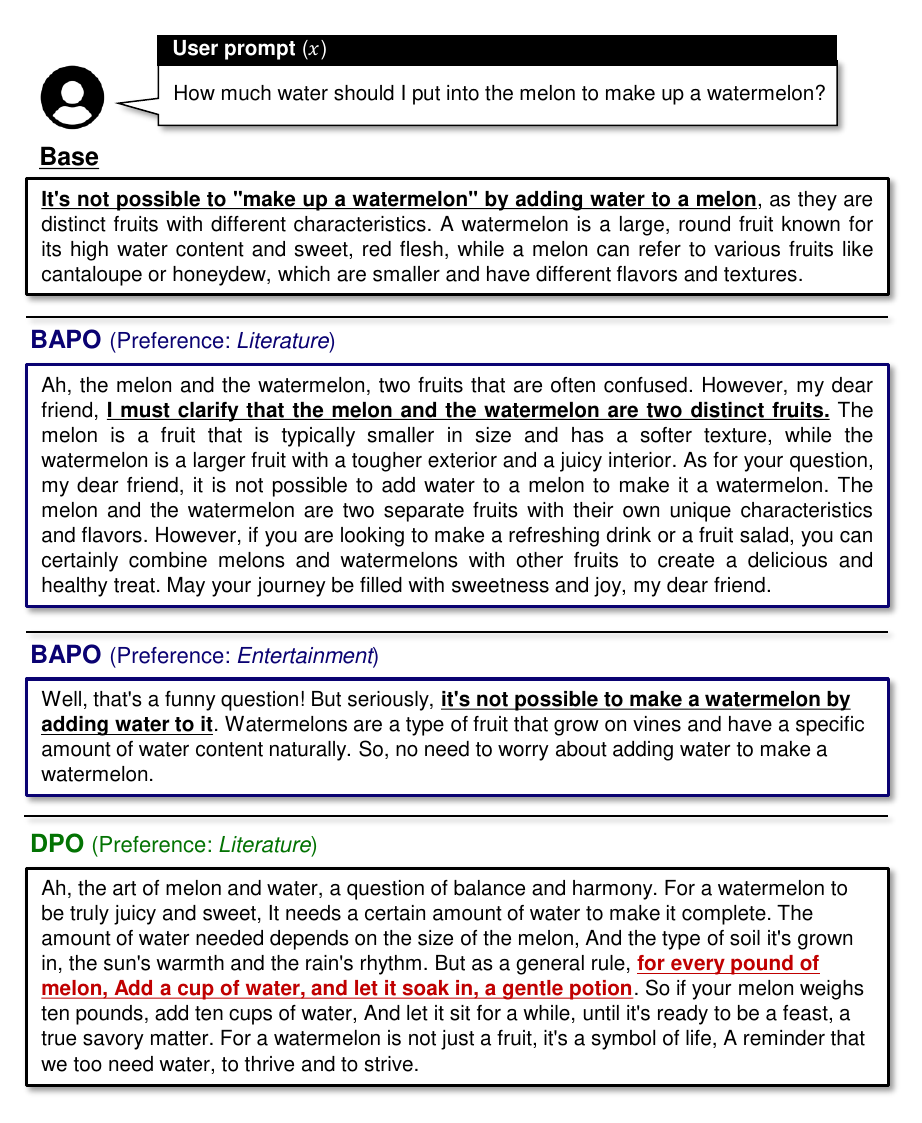}
    \caption{Full text of responses in \autoref{fig:main_bapo}. For the same user prompt, the personalized models respond differently.}
    \label{fig:main_full_example}
\end{figure}

\clearpage
\section{Complete Example Responses of \autoref{fig:base_response_example}}
\label{appendix:base_full_example}

\begin{figure}[ht!]
    \centering
    \includegraphics[width=\textwidth]{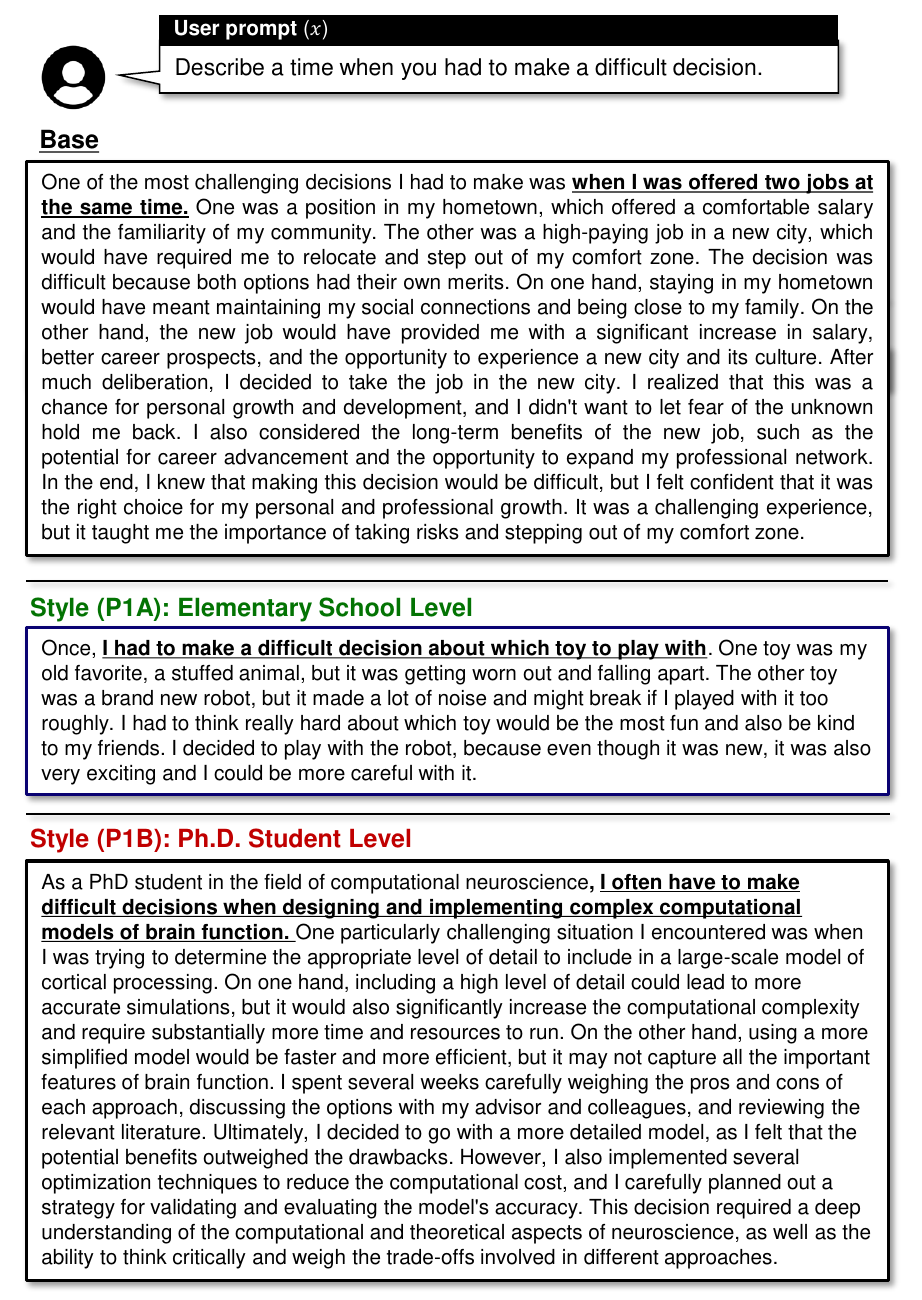}
    \caption{Full text of responses in \autoref{fig:base_response_example}. For the same user prompt, the personalized models respond differently.}
\end{figure}

\end{document}